\documentclass[journal]{IEEEtran}

\ifCLASSINFOpdf
  \usepackage[pdftex]{graphicx}
  \graphicspath{{./images}}
  \DeclareGraphicsExtensions{.pdf,.jpeg,.jpg,.png,.PDF,.JPEG,.PNG}

\usepackage{ifpdf}
\usepackage{cite}
\usepackage{xcolor}
\usepackage{amsmath}
\usepackage{amssymb}
\usepackage{amsthm}
\usepackage{amsfonts}
\usepackage{algorithm}
\usepackage{algorithmic}
\usepackage{multirow}
\usepackage{array}
\usepackage{booktabs}
\ifCLASSOPTIONcompsoc
 \usepackage[caption=false,font=normalsize,labelfont=sf,textfont=sf]{subfig}
\else
 \usepackage[caption=false,font=footnotesize]{subfig}

\usepackage{stfloats}

\usepackage{url}

\usepackage[switch]{lineno}

\begin{document}

\def\x{{\mathbf x}}
\def\L{{\cal L}}
\def\E{{\mathrm{E}}}

\def \TODO [#1]{\textcolor{red}{TODO: #1}}
\def \NOTE [#1]{\textcolor{blue}{(\textit{#1})}}
\def \comment [#1]{}
\def \etal {\emph{et. al.~}}

\def \check {$\checkmark$}
\def \cross {$\times$}

\def \sccaption [#1] {\caption{\\\textsc{#1}}}

\title{Towards Fine-grained Human Pose Transfer with Detail Replenishing Network}

%
%
%

\author{Lingbo~Yang, Pan~Wang, Chang~Liu, Zhanning~Gao, Peiran~Ren, Xinfeng Zhang, Shanshe~Wang,\\
        Siwei~Ma,~\IEEEmembership{Member,~IEEE,}
        Xiansheng~Hua,~\IEEEmembership{Fellow,~IEEE}
        and~Wen~Gao,~\IEEEmembership{Fellow,~IEEE}

\thanks{Lingbo Yang is a Ph. D. Student at Video Coding Laboratory, Institute of Digital Media, Peking University (PKU-IDM-VCL), Beijing, 100871 CHN. This work is done when he is interning at DAMO Academy, Alibaba, along with Pan Wang, Zhanning Gao, Peiran Ren, and Xiansheng Hua. Chang Liu and Xifeng Zhang are with University of Chinese Academy of Sciences. Shanshe Wang, Siwei Ma, and Wen Gao are with Institute of Digital Media, Peking University. Siwei Ma is the corresponding author.}%
\thanks{This work was supported by the National Natural Science Foundation of China (61632001), and High-performance Computing Platform of Peking University, which are gratefully acknowledged.}%
}

\maketitle


\begin{abstract}
Human pose transfer (HPT) is an emerging research topic with huge potential in fashion design, media production, online advertising and virtual reality. For these applications, the visual realism of fine-grained appearance details is crucial for production quality and user engagement. However, existing HPT methods often suffer from three fundamental issues: detail deficiency, content ambiguity and style inconsistency, which severely degrade the visual quality and realism of generated images.
Aiming towards real-world applications, we develop a more challenging yet practical HPT setting, termed as Fine-grained Human Pose Transfer (FHPT), with a higher focus on semantic fidelity and detail replenishment.
Concretely, we analyze the potential design flaws of existing methods via an illustrative example, and establish the core FHPT methodology by combing the idea of content synthesis and feature transfer together in a mutually-guided fashion. Thereafter, we substantiate the proposed methodology with a Detail Replenishing Network (DRN) and a corresponding coarse-to-fine model training scheme.
Moreover, we build up a complete suite of fine-grained evaluation protocols to address the challenges of FHPT in a comprehensive manner, including semantic analysis, structural detection and perceptual quality assessment. Extensive experiments on the DeepFashion benchmark dataset have verified the power of proposed benchmark against start-of-the-art works, with 12\%-14\% gain on top-10 retrieval recall, 5\% higher joint localization accuracy, and near 40\% gain on face identity preservation. Our codes, models and evaluation tools will be released at \url{https://github.com/Lotayou/RATE}
\end{abstract}


\ifCLASSOPTIONpeerreview
 \begin{center} \bfseries EDICS Category: 3-BBND \end{center}
\fi
%
\IEEEpeerreviewmaketitle

\section{Introduction}\label{sec:intro}

\IEEEPARstart{H}{uman} pose transfer (HPT) is an emerging research topic that attracts increasing attention recently. Aiming at synthesizing person images under new target poses with respect to the appearance of a given source image, HPT contains huge potential in empowering numerous creative applications, such as automatic fashion design, creative media production, online advertising and virtual reality. For these applications, the users would most likely focus their attention at semantically meaningful and detail-rich regions, such as face and clothes. Therefore, the ability to preserve semantic information and replenishing fine-grained appearance details is crucial for the performance and user experience of an HPT model.

\begin{figure*}[!t]
	\centering
    \includegraphics[width=\linewidth]{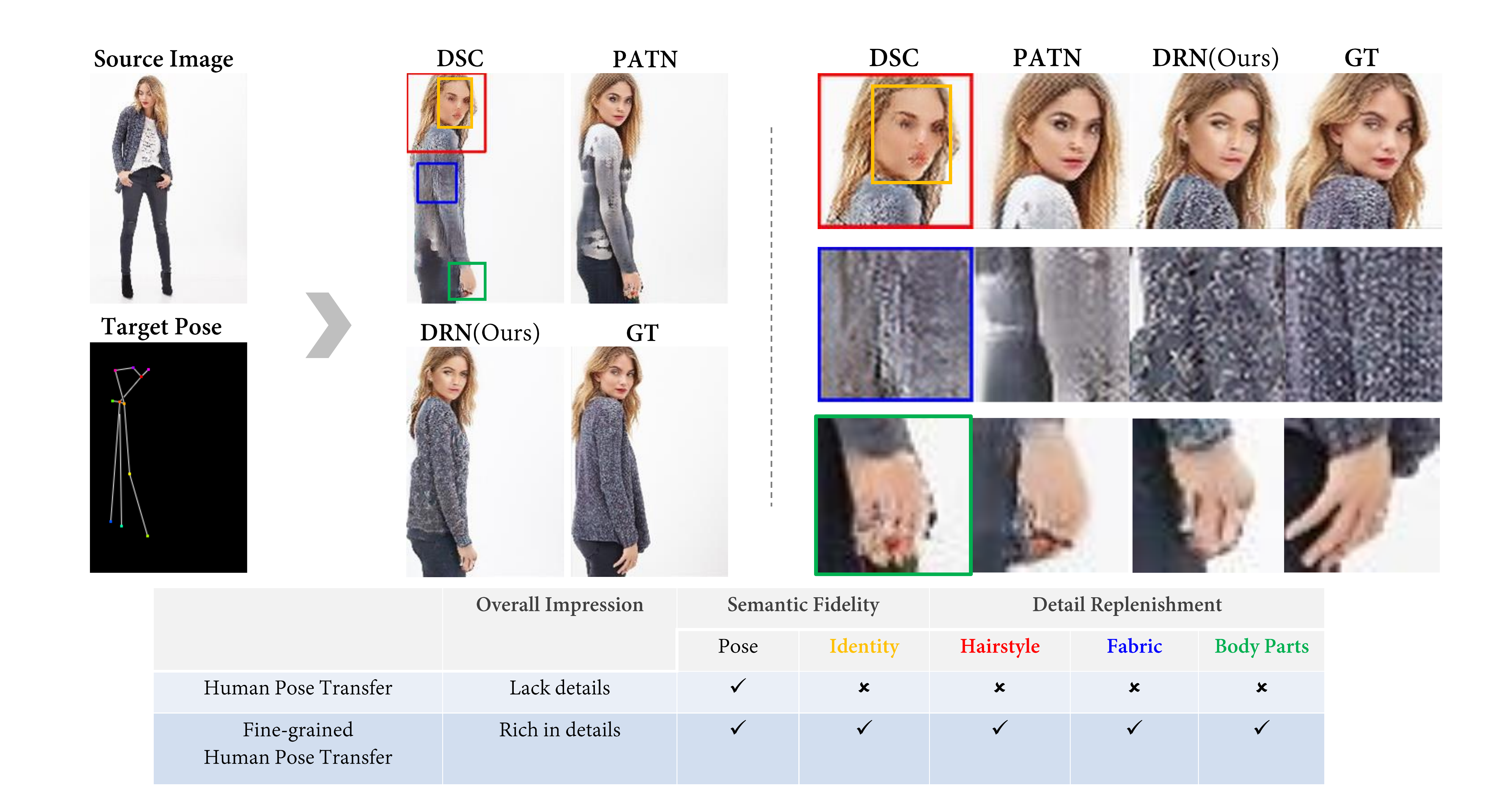}

	\caption[stunner]{A qualitative illustration of the FHPT objectives achievable by our method and state-of-the-art HPT methods: DSC~\cite{DeformableGAN} and PATN~\cite{PATN}. The detailed regions in colored bounding boxes are enlarged on the right. \emph{Please zoom in for details.}}
	\label{fig:stunner}
\end{figure*}

Inspired by the image-to-image translation~\cite{pix2pix}, early HPT solutions~\cite{PG2}\cite{disentangled}\cite{UVNet} adopt a global predictive strategy to directly translate the source image onto target poses by employing the U-net architecture with skip connections to propagate low-level features. However, due to the lack of localized deformation modeling, U-net based global methods often fail to properly handle the misalignment between source and target poses, leading to \emph{detail deficiency} in synthesized person images, such as over-smoothed clothes or distorted faces. Instead, Siarohin~\etal proposes a modified feature fusion mechanism~\cite{DeformableGAN} for warping the appearance features extracted from the source image onto the target pose with part-wise affine transformation, opening up a local warping approach for fine-grained appearance transfer. Thereafter, extensive research efforts have been dedicated to better modeling of body deformation and local feature transfer, including thin-plate spline function~\cite{softgate}, local attention~\cite{PATN}\cite{global_flow_local_attn}, optical flow~\cite{vid2vid}\cite{unsupervisedPF} and 3D surface models~\cite{DIAF}\cite{DensePoseTransfer}\cite{coordinate-based}.
However, as relying on exact feature correspondences for detail reconstruction, local warping methods typically suffer from content ambiguity in invisible regions where frequent viewpoint variations and occlusions occur in real-world applications.
To simultaneously overcome detail deficiency and content ambiguity, hybrid methods have emerged with an attempt to replenish new content with an additional predictive branch~\cite{vid2vid}\cite{fewshotvid2vid}\cite{liquidwarpinggan}. Yet the replenished contents often exhibit inferior perceptual quality than local-warped contents, incurring \emph{style inconsistency} in blended images.
Apparently, regarding real-world HPT applications, it is crucial to replenish fine-grained appearance details in a style-consistent manner for increasing performance and user experience.


To narrow the theory-practice gap of HPT, we aim towards a more practical and challenging setting of HPT, termed as Fine-grained Human Pose Transfer (FHPT). Specifically, FHPT addresses the aforementioned issues on detail deficiency, content ambiguity, and style inconsistency by emphasizing more on the preservation and replenishment of fine-grained semantic and appearance details, including facial identity, hairstyle, cloth fabrics, and small body parts, Fig.~\ref{fig:stunner}.
To implement FHPT, we propose the Detail Replenishing Network (DRN) with two distinctive designs: a style-guided detail replenishment module to enforce the style consistency of generated contents across the entire human body, and an intermediate feature-sharing path to facilitate the mutual guidance between the global predictive and local warping branch. Moreover, we establish a comprehensive suite of fine-grained evaluation protocols for more reliable and accurate measurement of the model capability towards FHPT objectives, including face identity preservation, keypoint localization, and content-based image retrieval.
Extensive experiments carried on the DeepFashion~\cite{DeepFashion} dataset verify the efficacy of our proposed DRN in preserving semantic attributes in the source image, as well as replenishing fine-grained appearance details in a style-consistent fashion. Compared to existing baselines, our method achieves 12\%-14\% gain on top-10 retrieval recall, 5\% higher joint localization accuracy, and near 40\% gain on face identity preservation, establishing a strong baseline method for FHPT.

This manuscript extends upon our previous work~\cite{MyICME2020} in three aspects. Firstly, we formally develop FHPT, a more practical and challenging HPT scenario with a higher emphasis on fine-grained semantic fidelity and detail quality.
Secondly, we propose the DRN for fine-grained person image generation and additionally address the limitation in identity preservation of our previous model by incorporating a facial attribute transfer module, leading to near 40\% higher facial identity preservation in synthesized images.
Thirdly, we develop a complete suite of fine-grained evaluation protocols for FHPT to better
analyze the efficacy of existing HPT techniques.

\section{Overview of Human Pose Transfer}\label{sec:FHPT_design}

As an emerging topic in computer vision, HPT is still in its infancy, where the first relevant research~\cite{PG2} was not proposed until 2017.
Driven by different research ideas in various related fields, including conditional GANs, style transfer~\cite{PerceptualLoss}, human pose estimation~\cite{Openpose}, and computer graphics~\cite{SMPL}, HPT has been evolving in a multifaceted manner. Yet there is still no conclusive answer to the rationality of these ideas, and their contribution towards the HPT objectives. Furthermore, most HPT works directly adopt existing evaluation metrics from related research fields, with limited ability in addressing the multifaceted nature of the HPT problem. In retrospect, the area of HPT would benefit from a more detailed and comprehensive objective design, as well as the corresponding evaluation protocols and research methodology towards practical applications. Here we provide a brief overview of the existing HPT methods and evaluation criteria, highlighting their potential limitations for preservation and replenishment of fine-grained semantic and appearance details.

\begin{figure*}[!pt]
	\centering
	\includegraphics[width=\linewidth]{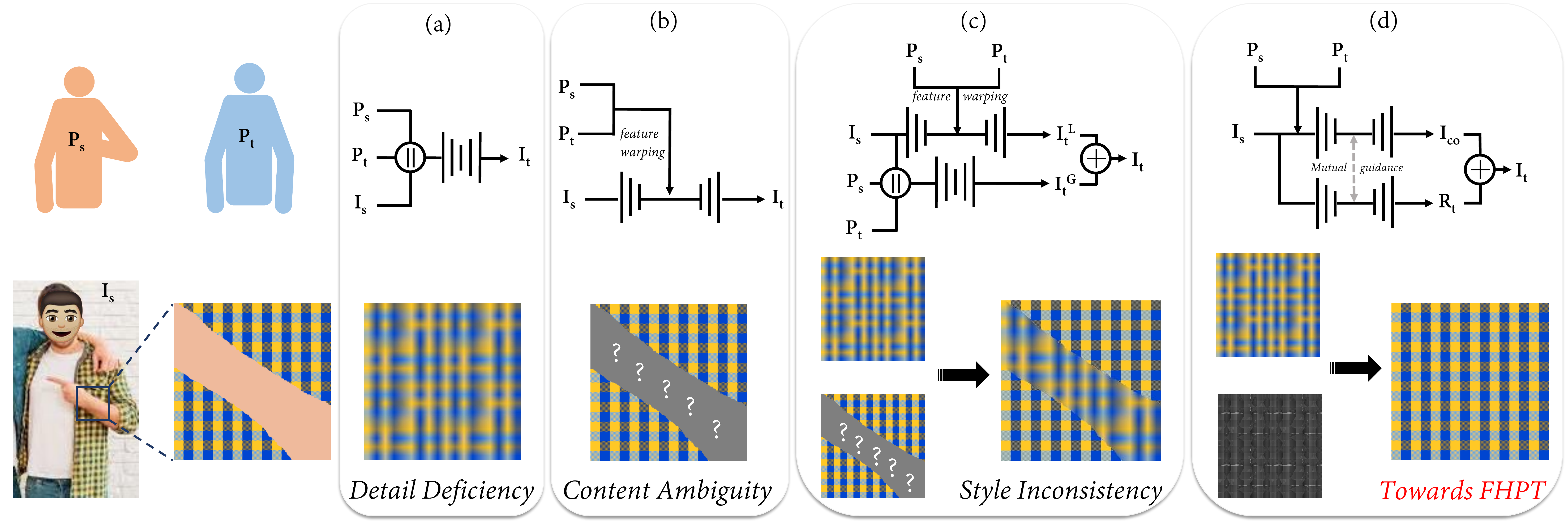}
	\caption{Illustration of different HPT network architectures and their potential design flaw. (a) Global predictive method --- detail deficiency; (b) Local warping method --- content ambiguity; (c) Hybrid method with independent branches --- style inconsistency; (d) Our proposed detail replenishing network conducts better feature utilization and mutual guidance with a feature-sharing path between two branches, which in consequence generates results with superior quality.
}\label{fig:architectures}
\end{figure*}

\subsection{HPT Methods}
Based on the feature transfer mechanism, we can summarize existing HPT methods into three categories: global predictive methods, local warping methods, and hybrid methods. Here we illustrate the behavior of different methods with a typical example in Fig.~\ref{fig:architectures}, where the source image contains a person in a blue-yellow checkerboard shirt partially occluded by his arm, and the target pose expects the person to move his arm away.
By speculating the output of different methods, we can reveal their potential limitations in two aspects,  including the ability to preserve semantic and appearance details in the source image and the ability to synthesize new content in occluded regions without exact feature correspondences.

\textbf{Global predictive methods}~\cite{PG2}\cite{disentangled}\cite{UVNet} typically formulate the HPT as a multi-modal image-to-image translation problem~\cite{pix2pix} and utilize the U-net architecture with skip connections for feature propagation. The pose guidance is introduced by encoding joint locations into a spatial heatmap~\cite{PG2} and concatenating it along with the input source image. Specifically, as shown in Fig.~\ref{fig:architectures} (a), the source pose $P_s$ and target pose $P_t$ are encoded into spatial guidance maps and concatenated along with the source image $I_s$ to directly generate a prediction $\tilde{I_t}$ of the target image $I_t$. However, such works often cannot reliably cope with the structural misalignment between different poses~\cite{DeformableGAN} due to the lack of accurate deformation modeling. Alternatively, several works in human motion transfer~\cite{EverybodyDanceNow}\cite{vid2vid}\cite{MyICME2019} aim to directly translate pose features to video frames, but often suffers from limited generalization towards new persons due to the lack of appearance guidance.
Generally speaking, global predictive methods typically suffer from the inability to capture localized feature correspondences, which often leads to \emph{detail deficiency} in synthesized images, e.g. blurry details and distortion artifacts, Fig.~\ref{fig:architectures} (a).

\textbf{Local warping methods} take inspiration from the spatial transformer networks~\cite{SpatialTN} and incorporate deformation modeling into the feature propagation mechanism.
Deformable Skip-Connection~(DSC)~\cite{DeformableGAN} is proposed for warping local features through part-wise affine transformation which establishes a new HPT methodology that later inspires a great amount of works. Following DSC, Thin-Plate Splines~\cite{TPS} have further facilitated non-linear warping estimation~\cite{softgate}, and local attention mechanism has been incorporated for increased flexibility in deformation modeling~\cite{PATN}\cite{global_flow_local_attn}. Moreover, the rapid advancement in automated dense 3D annotation~\cite{densepose}\cite{HMR} has enabled pixel-level feature warping flow estimation, enabling fine-grained appearance transfer from source to target images~\cite{DIAF}\cite{DensePoseTransfer}\cite{coordinate-based}.

The deformation mapping of the human body between different poses is usually determined via interpolation over a finite set of keypoint correspondences. Specifically, denote the region occupied by the human body in the source/target image as $\Omega_s$ and $\Omega_t$, one can formulate the deformation as a continuous mapping $T_{st}: \Omega_s \rightarrow \Omega_t$ that carries keypoints in the source pose to the target pose, \emph{i.e.} $T_{st}(p_s) = p_t, \forall (p_s, p_t)\in (P_s, P_t)$. Usually such a mapping is not unique, which necessitates an additional regularization term that sometimes referred to as the ``bending energy''~\cite{TPS} to mitigate the distortion of warped contents. The resulted mapping is usually composed of part-wise affine transformations~\cite{DeformableGAN} or thin-plate splines~\cite{softgate}.
Afterward, the output image is reconstructed upon the image features encoded from the source image and warped by the estimated deformation, which can be denoted as $\tilde{I}_t = \mathrm{Rec}\circ T_{st}\circ \mathrm{Enc}(I_s)$.

However, due to the frequent viewpoint changes and self-occlusions, there is no guarantee that the estimated warping would cover the entire target human body, \emph{i.e.} $\Omega_t - T_{st}(\Omega_s) \neq \phi$. Hence it would be difficult for a local warping network to faithfully recover the underlying content without exact correspondence in the source image, leading to \emph{content ambiguity} in uncovered regions, Fig.~\ref{fig:architectures} (b).

\textbf{Hybrid methods}~\cite{DensePoseTransfer}\cite{liquidwarpinggan}\cite{unsupervisedPF}\cite{fewshotvid2vid} aim to hallucinate new contents in uncovered regions with another global predictive branch, and blend the global and local generation results according to an estimated composition mask. Such a design is a common practice for video frame prediction~\cite{vid2vid}\cite{fewshotvid2vid} with little motion between consecutive frames and convenient warping via optical flow~\cite{flownet2.0}.
Furthermore, Zheng~\etal\cite{unsupervisedPF} designed an unsupervised flow learning scheme to tackle the unpaired scenario.
However, existing hybrid methods typically focus on blending the generation results at \emph{image-level}, whereas the intermediate-level feature fusion is less explored. Specifically, the global predictive branch often works separately from the local warping branch, which could incur \emph{style inconsistency} between hallucinated and warped contents, Fig.~\ref{fig:architectures}(c).

Considering the requirements for real-world applications, it is extremely crucial to develop a more suitable approach for replenishing fine-grained appearance details in uncovered regions without exact feature correspondences. Specifically, the semantic fidelity between source and target images, as well as the style consistency across the entire human body, should be effectively maintained.

\subsection{HPT Evaluation Criteria}\label{sec:HPT_eval}

For HPT, the quality of generated person images naturally comprises multiple factors, including perceptual quality, structural integrity, and semantic fidelity. Therefore, it is crucial to have well-designed measures addressing these factors simultaneously for reliable and accurate judgment on the capability of HPT models. However, most evaluation metrics are designed towards specific objectives, with less discriminative power in other aspects. For instance, PSNR is designed for measuring signal-level fidelity, but not well-suited for evaluating perceptual quality~\cite{niqe}; Inception Score~\cite{InceptionScore} is designed for measuring images with a multi-modal distribution, and is less reliable for the HPT task containing only a single object category~\cite{ISnote}.
Therefore, existing works typically introduce auxiliary evaluation tasks to validate the performance in other aspects, such as person detection\cite{DeformableGAN}\cite{DensePoseTransfer}, keypoint localization\cite{PATN}\cite{EverybodyDanceNow} and attribute prediction~\cite{DIAF}. However, most auxiliary evaluation tasks focus on high-level semantic aggregation, with less discriminative ability on fine-grained details. Also, the corresponding solutions to these tasks are by design robust, or in other words, insensitive to the noises and artifacts in generated images. Several works~\cite{PG2}\cite{DeformableGAN}\cite{PATN} have utilized the generated images for data augmentation on person Re-ID, but over the low-resolution ($128\times64$) Market1501~\cite{Market1501} benchmark dataset, making it unreliable for measuring the visual quality of high-resolution and detail-rich images.
Therefore, it is vital to develop a complete suite of evaluation protocols to assess HPT models in a reliable, accurate, and comprehensive fashion, which we will elaborate in the next section.

\section{Towards Fine-grained Human Pose Transfer}

In this section, we develop the proposed FHPT in three aspects: objectives, evaluation criteria, and the corresponding methodology. Specifically, we establish a new set of FHPT objectives to address the requirements in practical application scenarios with user interactions. Based on the proposed objectives, we develop a complete suite of fine-grained evaluation protocols for a more comprehensive, reliable, and accurate measurement of the capability of FHPT models. To fulfill the perspective FHPT objectives, we provide a new detail replenishment methodology along with a corresponding benchmark solution that will be further detailed in Sec.~\ref{sec:DRN}.

\subsection{FHPT Objectives}

Considering the requirements for creative design applications with complex garments and accessories of diverse color, fabric, texture, FHPT aims to focus on better style representation and attention to pattern details, and satisfy the natural inclination of human perception towards semantically-meaningful and detail-rich contents. Different from HPT objectives with a bias towards the accuracy of pose transfer, the proposed FHPT further emphasizes the preservation and replenishment of fine-grained semantic and appearance details, including facial identity, hairstyle, cloth fabrics, patterns, and small body parts, Fig.~\ref{fig:stunner}. Specifically, the FHPT objectives comprise three aspects:

\begin{itemize}
  \item \textbf{Perceptual Realism} The generated images should look natural and appealing, with rich, convincing appearance details over the entire image. Specifically, for hybrid networks, the predicted and warped image contents should be consistent in style and quality.
  \item \textbf{Structural Integrity} The generated images should well fit the target pose without noticeable structural distortion, particularly for facial landmarks and small body parts.
  \item \textbf{Semantic Fidelity} The generated images should preserve all necessary semantic attributes in the source image that help determine the person's identity, including both the facial attributes and the clothing appearance, such as color, hairstyle, and fabric.

\end{itemize}

\subsection{FHPT Evaluation Criteria}\label{sec:eval}

To better inspire the design of network architectures, loss functions, and training schemes for the proposed FHPT and offer accurate and reliable measurement for the corresponding models, we establish a comprehensive suite of fine-grained evaluation protocols targeting the FHPT objectives: semantic fidelity, structural integrity, and perceptual realism. Below we detail the evaluation protocols for each objective.

\textbf{Perceptual Evaluation Protocols}
Existing works often adopt Structural Similarity(SSIM)~\cite{SSIM} and  Inception Score (IS)~\cite{InceptionScore} to account for the statistical and perceptual fidelity of generated images. However, a recent study~\cite{ISnote} has shown that the Inception Score is susceptible to network weights, batch size, and data distribution, making it unreliable for measuring the quality of generative models. To address this issue, we introduce two supervised perceptual metrics: FID~\cite{FID} and LPIPS~\cite{LPIPS}, to better reflect the perceptual quality of our framework. Both metrics utilize a pre-trained network to project images onto the feature space and compute the distance between image features for the entire distribution or individual pair of samples. Compared with the unsupervised IS metric, the FID and LPIPS metrics can better reflect the perceptual quality of generated images.

\textbf{Structural Evaluation Protocols}
We introduce a new metric called Keypoint Error Curve \emph(KEC) to evaluate the localization accuracy of small body parts in generated images. It is similar to the PCKh metric in pose estimation tasks~\cite{PATN}, but refined in two ways: (1) In addition to body joints, we also extract landmarks for the face and both hands, and evaluate the accuracy separately for each body part. This helps better reflect the capability of an FHPT model in maintaining the structural integrity of small body parts. (2) We provide a more detailed performance profile by evaluating the keypoint estimation accuracy at a set of adaptively selected anchors. For each given threshold $\alpha$, we calculate the percentage of detected keypoints $P(\alpha)$ with distance to the ground truth smaller than $\alpha$ and plot the accuracy curve.

\textbf{Semantic Evaluation Protocols}
As most semantic information in typical person images concentrates on facial attributes and clothing contents, we propose two well-defined and interpretable measurements for semantic fidelity: content-based image retrieval and face identification. For retrieval, we utilize generated images to query from the database of all corresponding source images, and calculate the retrieval scores using ground truth annotations provided in~\cite{DeepFashion}. Specifically, the retrieval-based evaluation metric is superior to existing distance metrics in three aspects:

\begin{enumerate}
    \item The retrieval task is more application-driven with customers being the final judge of content similarity. Thus a well-trained retrieval system will work effectively in extracting the most informative and discriminative features closely related to human perception.
    \item The retrieval-based evaluation protocol does not require the ground truth to be under the same pose as the query, making it more practical for real application purposes. Furthermore, its robustness against pose and view variations also indicates a better focus on semantic and appearance details.
    \item The reliability of retrieval-based metrics can be easily qualified by querying with real images, and further improved by fine-tuning the retrieval system over the testing dataset. In contrast, the performance bound for existing perceptual metrics is always the same, 0 for LPIPS and 1 for SSIM, making it difficult to quantify and improve the reliability of such metrics.
\end{enumerate}

To measure the consistency of face identity across source/target image pairs, we extract feature embeddings of cropped facial regions with a pre-trained face recognition model~\cite{dlib_paper}, and compare the embedding distance between each pair. If the distance is lower than a threshold $\epsilon$, the person's identity is considered to be preserved. In practice, we choose $\epsilon$ to be $0.6$ and $0.7$.

\subsection{FHPT Methodology}

\begin{table}[!pt]
	\renewcommand{\arraystretch}{1.2}
	\centering
	\caption{Main capabilities of different models, (a) Global predictive methods, (b) Local warping methods, (c) Hybrid methods, (d) Our DRN.}
	\label{tab:architectures}
	\begin{tabular}{|c|cccc|}
		\hline
		Networks & (a) & (b) & (c) & (d) \\
		\hline
        Explicit deformation modeling   & \cross & \check & \check & \check\\
        Pose-guided appearance transfer   & \cross & \check & \check & \check\\
        Multi-branch architecture & \cross & \cross & \check & \check\\
        Mutual guidance across branches & \cross & \cross & \cross & \check\\
        Fine-grained detail synthesis    & \cross & \cross & \cross & \check\\
		\hline
	\end{tabular}
\end{table}

So far, we have discussed about existing HPT approaches in terms of feature utilization and content generalization schemes, and highlight their potential flaws in addressing the FHPT challenges. Specifically, a fundamental problem in FHPT lies in the preservation and replenishment of fine-grained semantic and appearance details, which have been explored from two separate approaches. One is to directly synthesize new content based on high-level semantic guidance as in global predictive methods~\cite{disentangled}\cite{adgan}, and the other is to transfer low-level visual features from the source image, featuring local warping methods with warping flow~\cite{vid2vid} or attention mechanism~\cite{PATN}. Consequently, the task of human pose transfer is also known by different names, such as ``pose-guided human (person) image generation (synthesis)'', indicating the methodological divergence in existing research efforts. Needless to say, both approaches have their advantages and limitations, which necessitates an effective integration scheme to maximize the complementary effects between two approaches. With the emergence of hybrid architectures, the two opposite ends start to join together, but the integration result is still sub-optimal mainly due to the inconsistency between globally and locally generated content.

Moving towards FHPT, the transfer of available image contents and the synthesis of new content should be carried out in a mutually-guided fashion to simultaneously promote the style consistency and detail quality of the final output. To this end, we propose the Detail Replenishing Network (DRN) to substantiate the FHPT methodology with two distinctive architectural designs: 1) a guided detail replenishing module to enforce style-consistency; 2) an intermediate feature sharing pathway to promote mutual guidance.
Table~\ref{tab:architectures} summarizes the main capabilities of DRN against existing networks in Fig.~\ref{fig:architectures}(a)-(c), where DRN works in a ``coarse-to-fine'' fashion by synthesizing a global detail replenishing residual map $R_t$ upon the coarse estimation $\tilde{I}_c$, leading to \emph{detail-rich}, \emph{unambiguous} and \emph{style-consistent} generation results. In the next section, we will detail the implementation of individual components of the conceptualized FHPT model in Fig.~\ref{fig:architectures}(d), and introduce the corresponding model learning scheme to better fulfill the proposed FHPT objectives.

\begin{figure*}[!t]
	\centering
	\includegraphics[width=\linewidth]{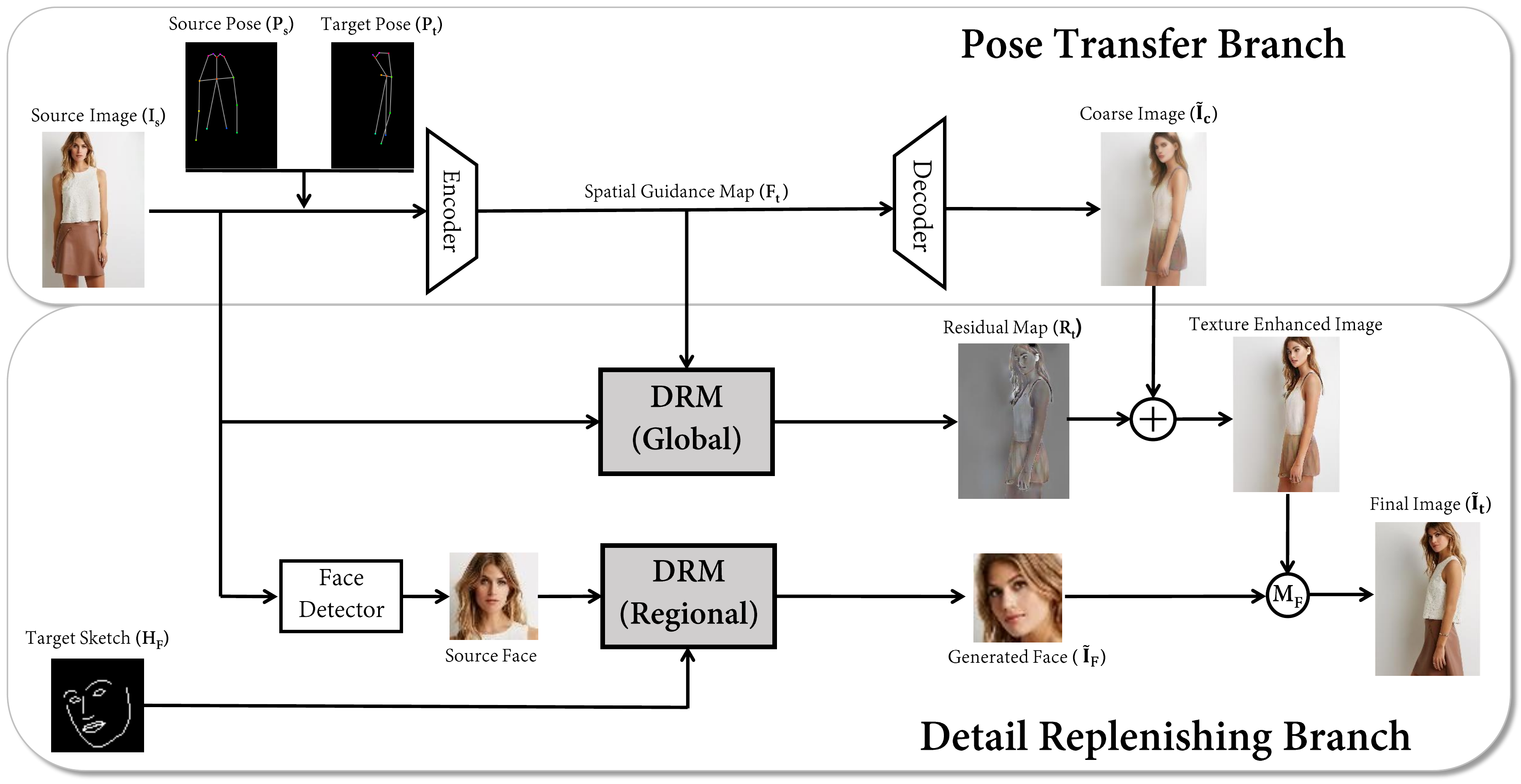}
	\caption{Overview of the proposed DRN. The pose transfer branch first estimates a coarse output $\tilde{I}_{co}$ under the target pose. The content feature map $F_t$ provides spatial guidance to the detail replenishing branch, which then generates a residual map $R_t$ to refine the coarse output. Also, the detail replenishing branch also helps preserve the person's identity by generating the face under target view given source face (if visible) and target sketch $H_F$. 
}\label{fig:framework}
\end{figure*}

\section{Detail Replenishing Network}\label{sec:DRN}

In this section, we illustrate the architectural design and training details of the proposed Detail Replenishing Network (DRN) which contains two branches: A pose transfer branch for acquiring a coarse estimation and provide spatial guidance, and a detail replenishing branch for refining local visual details in a style-guided fashion, Fig.~\ref{fig:framework}.

\subsection{Network Architecture}\label{section:pose}
\textbf{Pose Representation}
The human pose representation can be implemented in various ways regarding the resource constraints and application considerations. In this paper, we choose a computationally efficient pose representation of sparse 2D keypoints. For each image $I$ in the training dataset, we extract an 18-point pose skeleton $P$ using a pretrained estimation network~\cite{Openpose}, and convert the skeleton coordinates into an 18-channel pose heatmap. Dense representations, such as body parsing labels~\cite{softgate}, optical flows~\cite{fewshotvid2vid} or pseudo-3D surfaces~\cite{DensePoseTransfer} are also compatible with our framework at the expense of computational cost and manipulation inflexibility. Furthermore, we also extract the face bounding box $B$ using a lightweight face detection library~\cite{dlib_paper} to help preserve necessary facial attributes. Also, the landmarks of the target face are estimated with~\cite{Openpose} and encoded as a sketch $H_F$, which will be provided during inference if the front view is available in the target image.

\textbf{Pose Transfer Branch} takes in the source image $I_s$ and the paired pose representation~$(P_s, P_t)$, and aims to generate a coarse estimation of the target image. The network begins with several down-sampling convolutional layers to encode the contents of the source image $I_s$ and the pose dependencies between $P_s$ and $P_t$. To better capture the relationship between different poses, the pose pair $(P_s, P_t)$ is concatenated along the depth axis before sent into the encoder. We leverage the design in PATN~\cite{PATN} by introducing several cascaded transfer blocks to encourage a smooth transition of encoded contents. The result feature map $F_t$ is roughly aligned with the target pose, which will serve as the spatial guidance for detail replenishment (see Sec. for visualization results).
Finally, we decode $F_t$ with several upsampling layers to acquire a coarse estimation of the target image.
\begin{equation}
\tilde{I}_{c} = G_P(I_s, P_s, P_t)\label{eqn:coarse}
\end{equation}

Note that for our proposed framework, it is crucial that the transfer branch provide accurate spatial guidance for the upcoming detail replenishing branch. In particular, we \emph{do not} want any fake details to be injected into the guidance map $F_t$ and the coarse estimation $\tilde{I}_{c}$. Instead, the transfer branch is supposed to provide useful hints on \emph{``where to add what kind of details''} and let the detail replenishing branch do the work. This allows a coarse-to-fine generation of the final image with better visual quality. Ablation studies are provided in section~\ref{sec:exp} to justify our claims.

\textbf{Detail Replenishing Branch} is composed of several detail-replenishing modules (DRM) that each specializes in refining the entire image or a specific region. For our benchmark solution, we adopt both types of DRMs to refine both the entire image and the face region.
The global module predicts a residual map $R_t$ from the spatial guidance map $F_t$ and the source image $I_s$, and add the residual map onto the coarse result $I_c$; while the regional (face) module directly synthesizes the target face image $I_F$ under the target landmark sketch $H_F$ according to the facial attributes extracted from the input face crop. In this way, the face module can be trained independently from the global module and can benefit from additional face data. To assemble the final output, we perform an alpha blending with Gaussian blurred weight mask $M_F = \mathrm{g}(\sigma) * \mathbf{1}_{B_F}$, where $*$ denotes the 2D convolution operator, $\mathrm{g}(\sigma)$ is a discrete 2D Gaussian kernel, and $\mathbf{1}_{B_F}$ is an indicator function that equals $1$ if the corresponding pixel belongs to the face bounding box $B_F$ and $0$ elsewhere. Incidentally, we only need to activate the face enhancing module if the frontal face is available in both the source and the target image.
Finally, combining the results in Eqn.~\eqref{eqn:coarse} lead to the final result:
\begin{equation}
\tilde{I}_t = (\mathbf{1} - M_F)(\tilde{I}_{c} + R_t) + M_F\tilde{I}_F\label{eqn:final}
\end{equation}

\begin{figure}[!t]
	\centering
	\includegraphics[width=\linewidth]{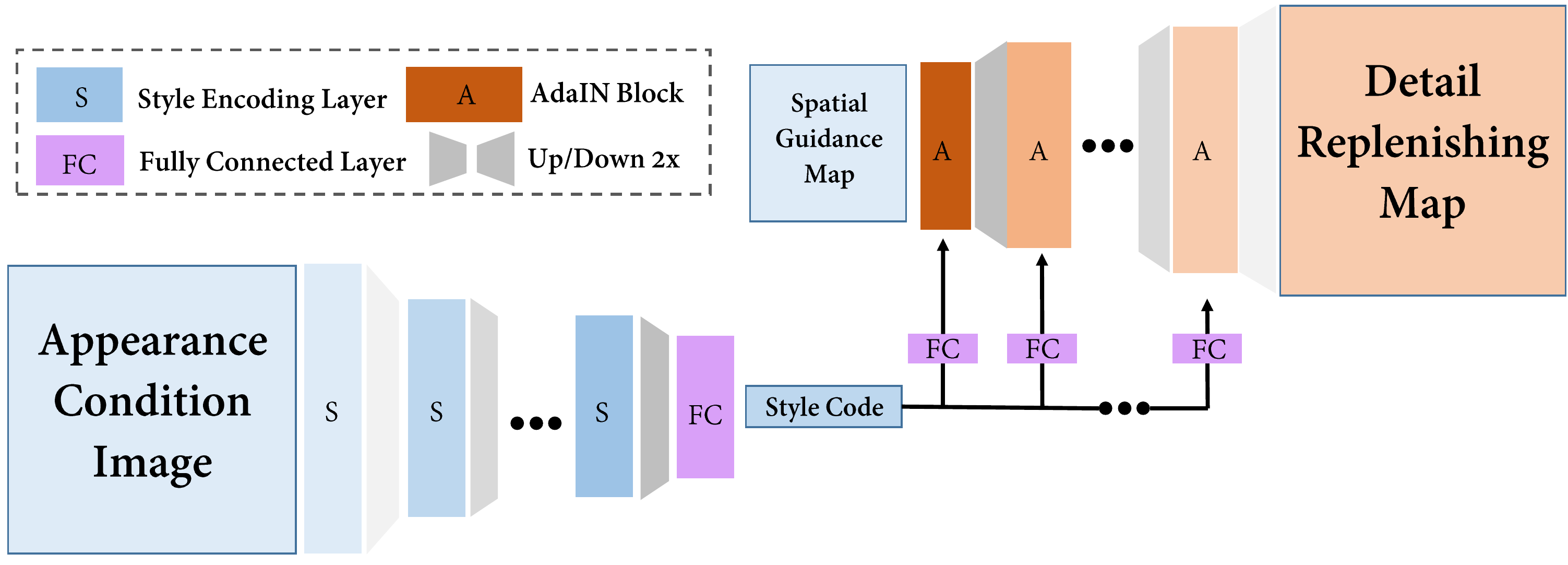}
	\caption{Network architecture of the proposed detail replenishment module.}\label{fig:drum}
\end{figure}

\textbf{Detail Replenishment Module}\label{sec:drm}
The proposed detail replenishment module~(DRM) aims to encode the visual attributes of the appearance condition image $I_s$ and generate an detail replenishing map~$R_t$ following the spatial guidance map $F_t$. The network architecture of the DRM is shown in Fig.~\ref{fig:drum}.

We extract the appearance and style information from the source image $I_s$ with an encoder consisting of several style-encoding residual blocks. To acquire a more robust representation against partially occluded clothes, we perform global average pooling to convert the encoded feature tensor into a style code $z_s$. After the style code $z_s$ is acquired, an enhancing residual map $R_t$ (or target face crop $\tilde{I}_F$) is generated with respect to both the spatial and appearance guidance:

\begin{equation}
R_t = G_T(z_s,F_t)\label{eqn:res}
\end{equation}

where $F_t$ is the spatial guidance map described in section~\ref{section:pose}. Inspired by related works in style-based image generation~\cite{stylegan}, we utilize the adaptive instance normalization (AdaIN)~\cite{adain} to infuse the code into the output enhancing map. Note that $z_s$ now assumes the role of ``style-guide'' by controlling the pattern and granularity of synthesized details in different regions. The residual map generator $G_T$ is formed by several up-sampling convolutional layers. The content feature map $F_t$ at the coarsest scale is fed into the generator, where semantic and appearance details are gradually infused in a coarse-to-fine fashion across multiple layers. At the i-th layer, the texture code is processed with a 3-layer MLP to acquire the modulation weight and bias $\mathbf{s}^i = (s_w^i,s_b^i)$, which is then used for controlling the AdaIN operation:

\begin{equation}
\mathrm{AdaIN}(F_t^i, \mathbf{s}^i) = s_w^i \frac{F_t^i - \mu_t^i}{\sigma_t^i} + s_b^i
\end{equation}

where $\mu_t^i$ and $\sigma_t^i$ are the pre-channel mean and variance of the feature map $F_t^i$, the input at the $i$-th AdaIN layer.

\begin{algorithm}[!t]
\caption{Alternate optimization algorithm for the proposed network.}\label{alt_train}
\begin{algorithmic}
\REQUIRE Networks $G_P, G_T, D_A, D_S$; Dataloader $L$
\REPEAT
\STATE $I_s, I_t, P_s, P_t \leftarrow L.{next\_batch()} $
\STATE $\tilde{I_{c}}, F_t \leftarrow G_P(I_s, P_s, P_t)$
\STATE $L_1 \leftarrow \L_1(\tilde{I_{c}}, I_t)$
\STATE \textbf{Backward $L_1$}
\STATE \textbf{Update $G_P$}
\STATE $L_1 \leftarrow \L_1(\tilde{I_{c}}, I_t)$
\STATE $\tilde{I_{t}} \leftarrow \tilde{I_{c}} + G_T(I_s, F_t)$
\STATE $L_2 \leftarrow \L_2(\tilde{I_{t}}, I_t)$
\STATE \textbf{Backward $L_2$}
\STATE \textbf{Update $G_P$ and $G_T$}
\FOR {$i=1$ \textbf{in} $K$}
\STATE $I_s, I_t, P_s, P_t \leftarrow L.{next\_batch()} $
\STATE $\tilde{I_{c}}, F_t \leftarrow G_P(I_s, P_s, P_t)$
\STATE $\tilde{I_{t}} \leftarrow \tilde{I_{c}} + G_T(I_s, F_t)$
\STATE $R_S = D_S(\tilde{I}_t; P_t);\qquad R_A = D_A(\tilde{I}_t; I_S)$
\STATE $L_{GAN} = \L_{GAN}(R_S, R_A)$
\STATE \textbf{Backward $L_{GAN}$}
\STATE \textbf{Update $D_A$ and $D_S$}
\ENDFOR
\UNTIL {Convergence.}
\end{algorithmic}
\end{algorithm}

\subsection{Training Losses}
We adopt slightly different loss functions for two branches. For the transfer branch, the output coarse estimation $\tilde{I}_c$ is \textbf{not} expected to contain any fake textures, as explained in section~\ref{section:pose}. Therefore, we adopt the following loss formulation $\L_1$ to update the transfer branch without adversarial losses:

\begin{equation}
\L_1 = \lambda_{recon}\L_{recon} + \lambda_{per}\L_{per}\label{eqn:l1}
\end{equation}
where $\L_{recon}$ is the pixelwise L1 loss, and $\L_{per}$ is the perceptual loss in~\cite{PerceptualLoss}:

\begin{equation}
\L_{per} = \frac{1}{CHW}\sum_{l}\|\phi_l(I_t) - \phi_l(\tilde{I}_{co})\|
\end{equation}
Here $\phi$ is a pretrained VGG-19 network and $l$ denotes the layer index.
For the enhancing branch, we further add style and adversarial loss terms upon $\L_1$ to promote visual details, leading to the full loss function as follows:

\begin{equation}
\L_2 = \lambda_{recon}\L_{recon} + \lambda_{per}\L_{per} + \lambda_{sty}\L_{sty} + \lambda_{GAN}\L_{GAN}\label{eqn:l2}
\end{equation}

where $\L_{sty}$ is the Gram-matrix based style loss~\cite{PerceptualLoss}:
\begin{equation}
\L_{sty} = \frac{1}{CHW}\sum_l\|G(\phi_l(I_t)) - G(\phi_l(\tilde{I}_t))\|_F^2
\end{equation}

where $G$ is the Gram matrix:
$$ G(F)_{ij} = \frac{1}{CHW}\sum_{h=1}^{H}\sum_{w=1}^W F_{ihw}F_{jhw} $$

The adversarial loss $\L_{GAN}$ is evaluated with two conditional discriminators, $D_A$ and $D_P$ to measure the appearance and pose consistency respectively. Concretely, the appearance discriminator $D_A$ takes a pair of source/target images as input and tries to distinguish between real pairs $(I_s, I_t)$ and fake pairs $(I_S, \tilde{I}_t)$. Similarly, the pose discriminator $D_S$ takes in image/pose pairs~$(\tilde{I}_t; P_t)$ and evaluates if the generated images are consistent with the given pose conditions. Combing the shape and appearance loss leads to the following formulation:

\begin{multline*}
\L_{GAN} = \E(\log(D_A(I_s;\tilde{I}_t))) + \E(\log(1-D_A(I_s;I_t))) \\ + \E(\log(D_S(P_t;\tilde{I}_t))) + \E(\log(1-D_S(P_t;I_t)))
\end{multline*}

which is also used for discriminator updating. In practice, we use the LSGAN variant~\cite{LSGAN} for better training stability. For face enhancement, the landmark sketch $H_F$ is adopted as pose condition to replace the global pose $P_t$, and the source image $I_s$ is replaced with the cropped face patch, naturally.

\subsection{Optimization Algorithm}
We employ a coarse-to-fine gradient descent optimization algorithm to update the components alternatively. For each iteration, we first update the pose transfer branch with loss function $\L_1$ over the coarse estimation result $\tilde{I}_{c}$. This allows the spatial guidance map $F_t$ to be roughly aligned with target pose. Then we perform an end-to-end fine-tuning to update the enhancing branch with the detail-aware loss $\L_2$ computed over final output $\tilde{I}_t$
\footnote{The face enhancing module should be optimized separately, as it operates on cropped face regions instead of the entire image and the facial contents are in principle independent of other body parts.}.
Note that the pose transfer branch will also be fine-tuned in this step, as gradients can be backpropagated through $F_t$ and the preceding pose transfer blocks. The discriminators are then updated for $K$ steps, where we empirically choose $K=3$ to balance between running speed and discriminative capability. The complete training routine is shown in Algorithm~\ref{alt_train}. Other model learning schemes are compared in ablation study in section \ref{sec:ablation}.

\section{Experiments}\label{sec:exp}

In this section, we evaluate the performance of the proposed DRN against several competitive baseline methods. Furthermore, we perform an ablation study to verify the efficacy of our main contributions.

\textbf{Datasets} We carry all experiments on the \emph{In-shop Clothes Retrieval Benchmark} of the DeepFashion dataset~\cite{DeepFashion}, which contains more than 50,000 editorial images of fashion models under varying poses with texture-rich garments. Following the pre-processing routine in~\cite{PATN}, we crop the background from both side of the image by 40 pixels, leaving the center 256 $\times$ 176 region for training and testing. For pose representations, we extract the body skeletons and facial landmarks using Openpose~\cite{Openpose} and detect the face bounding box using\cite{dlib_paper}. We use the data partition in~\cite{PATN} with 101,966 pairs for training and 8,570 pairs for testing. The partition ensures that the same person will not appear in both the training and testing split, thus can fully guarantee the generalization ability of our DRN.

\textbf{Implementation Details}
We use PyTorch to implement our proposed framework. The transfer branch contains 3 down-sampling layers and 9 cascaded pose transfer blocks~\cite{PATN}, and the enhancing branch contains 6 residual blocks~\cite{Resnet}. LeakyReLU is introduced after each normalization layer with 0.2 negative slope. The length of texture codes is fixed to 128, where the impact of different lengths are further analyzed in ablation study, see section~\ref{sec:ablation} for details. Rectified Adam optimizer~\cite{radam} is adopted with $1e-5$ weight regularization for improved training stability and better final performance. The full framework is updated for involves 40 batches with a total of 200K iterations. The learning rate is initialized to $1e-4$ for all modules and is kept fixed for the first 10 batches before linearly decaying to 0. The loss weights $(\lambda_{GAN}, \lambda_{per}, \lambda_{recon}, \lambda_{sty})$ are set to ($1.0,5.0,10.0,5.0$). For face enhancement, we retain the training samples with paired face bounding boxes, and resize the face crops to $64\times64$ using bicubic interpolation. The face loss is computed by Eqn.~\eqref{eqn:l2} with the same loss weights. The training continues for 200 epochs with the learning rate fixed to $lr=1e-4$.

\subsection{Ablation Study}\label{sec:ablation}
In this section, we compare the proposed DRN against four ablation methods to analyze the impact of different components. The first two aim to verify the proposed network architecture, while the other two focus on the proposed alternate training strategy. We will leave the ablation study on facial detail replenishment module in section~\ref{sec:face}, where a more targeted analysis will be provided on cropped facial regions. Here we detail the settings of each ablation method as follows:

\textbf{PB Only} We remove the enhancing branch and train the pose transfer branch directly in an end-to-end fashion with loss function $\L_2$. Notice that this setting differs from the PATN~\cite{PATN} baseline where an additional style loss term is introduced with slightly adjusted loss weights.

\textbf{Texture64dim} We reduce the length of the textural code to 64 and keep the rest unchanged. In this way, the appearance information extracted from source image is reduced, which could result in less amount of details and more severe artifacts.

\textbf{PB Fix} We initialize the parameters of the transfer branch using pre-trained models in~\cite{PATN}, and keep it fixed during training. In this way, the interaction between two branches is broken, and detailed style loss and conditional adversarial losses cannot be back-propagated into the transfer branch.

\textbf{End2end} We randomly initialize the parameters and start training from scratch. Instead of the proposed alternative training, we directly update the whole framework with $\L_2$ defined over the refined output $\tilde{I_t}$ without explicit constraints on the coarse estimation $\tilde{I}_{c}$. In this way, the detail replenishing branch has to rely on inaccurate spatial guidance maps, which could lead to difficulty in convergence and suboptimal performance.

\begin{table}[!t]
	\renewcommand{\arraystretch}{1.2}
	\centering
	\caption{Quantitative ablation study results, the best result under each metric is shown in bold format.}
	\label{tab:ablation}
	\begin{tabular}{|c|cccc|}
		\hline
		Methods & SSIM $ \uparrow $ & IS $ \uparrow $ & FID $ \downarrow $ & LPIPS $ \downarrow $ \\
		\hline
		PB Only          & 0.772 & \textbf{3.231} & 18.602 & 0.250 \\
		Texture64dim     & 0.763 & 3.040 & 18.263 & 0.243 \\
		PB Fixed         & 0.767 & 2.923 & 20.346 & 0.238 \\
		End2end          & 0.768 & 3.147 & 16.496 & 0.231\\
		Ours Full        & \textbf{0.774} & 3.125 & \textbf{14.611} & \textbf{0.218}\\
		\hline
	\end{tabular}
\end{table}

\begin{figure}[t]
	\centering
	\includegraphics[width=\linewidth]{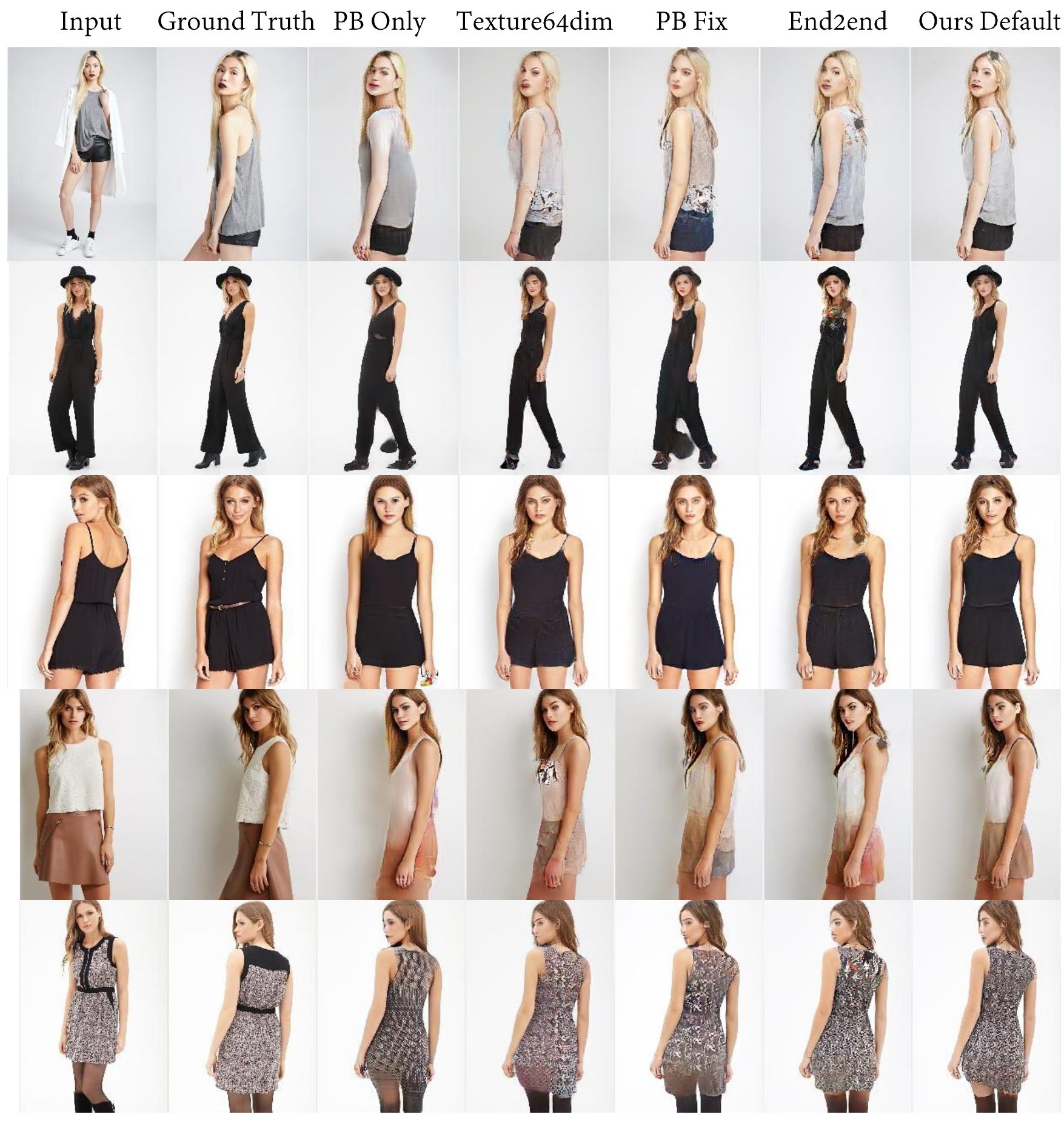}
	\caption{Qualitative results of ablation methods. Our full method faithfully capture the texture details as well as maintaining the integrity of body parts. \emph{Please zoom in for details.}}\label{fig:ablation}
\end{figure}

We present the scores evaluated on the DeepFashion dataset in Table~\ref{tab:ablation}. Our full framework, DRN, consistently outperforms other ablation methods, especially on perceptual metrics such as FID and LPIPS, highlighting the importance of detail replenishment for human image generation. Also we note that reducing the dimension of latent code leads to significant performance drop, indicating a potential trade-off between perceptual quality and memory cost.
Furthermore, Fig.~\ref{fig:ablation} showcases the quality improvement upon other ablation methods. The proposed DRN is capable of creating more authentic appearance details and suppressing repetitive artifacts. Also, due to the guidance from the texture-aware loss $\L_2$ over final outputs, it can better preserve the integrity of body parts, as observed in the second and the third row, indicating an increased accuracy of the estimated spatial guidance map $F_t$. Therefore, the effect of mutual guidance between both modules, along with the alternate optimization algorithm, is further justified.

\subsection{Comparisons with Previous Works}

We choose several representative state-of-the-art works as baselines: DSC~\cite{DeformableGAN}, UV-Net~\cite{UVNet}, SPT~\cite{SPT} and PATN~\cite{PATN}. Here DSC and PATN are retrained from scratch upon the same training/testing split as ours, where for DSC we have removed 426 bad cases with missing joints that would cause an error during inference, and report the scores on remaining samples. For UV-Net and SPT, the training script is either unavailable or not working as expected, so we report the scores in their original papers.

\begin{figure}[!t]
	\centering
	\includegraphics[width=\linewidth]{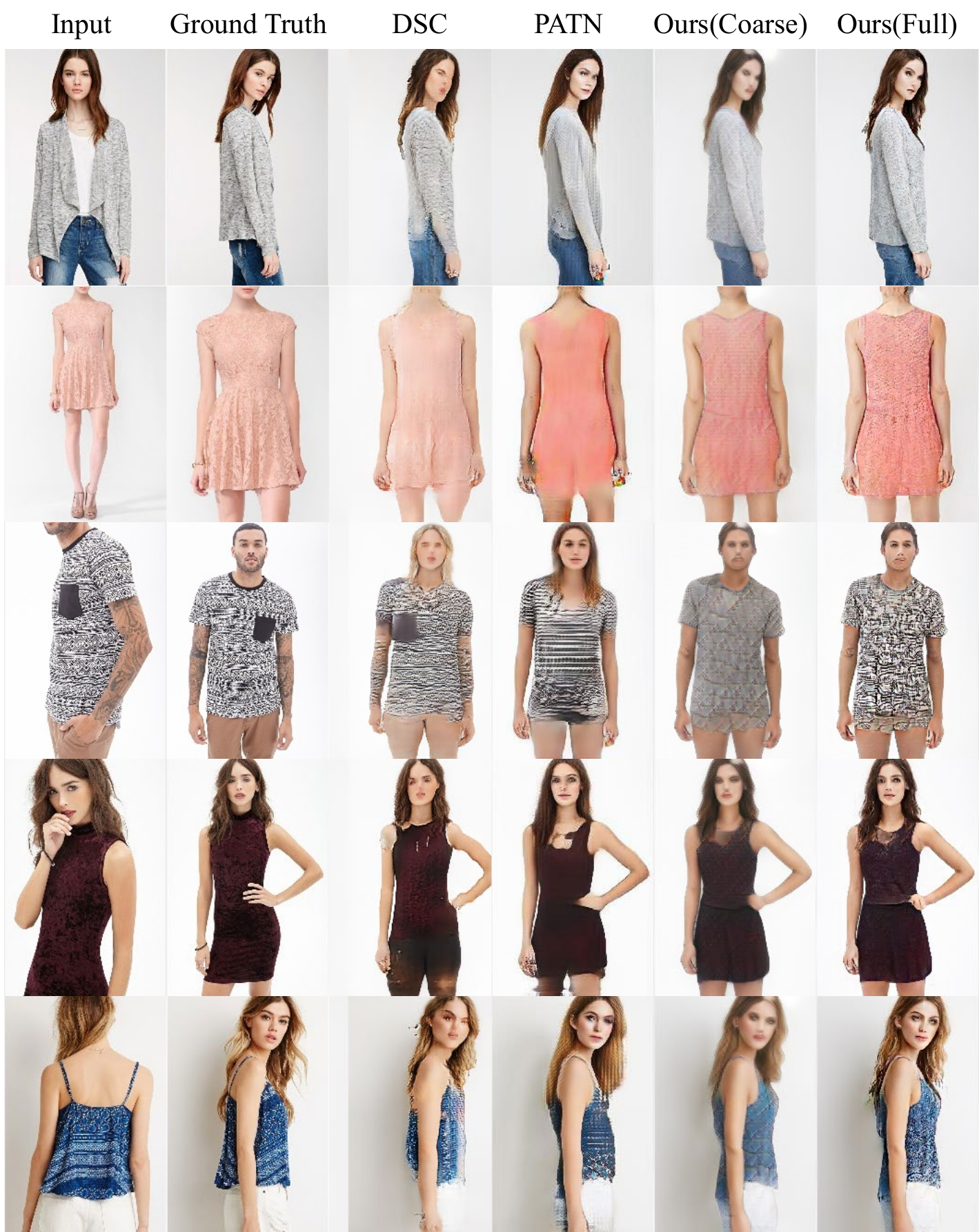}
	\caption{Qualitative results of the proposed method against several competitive baselines. Some images have been cropped for visualization purposes. \emph{Please zoom in for details.}}\label{fig:sota}
\end{figure}

\textbf{Qualitative Comparison}
Fig.~\ref{fig:sota} showcases the generation results for some challenging examples with large pose variations and rich textures. In general, our method is more faithful to the input conditions and contains more realistic semantic and appearance details, such as clothing fabrics (linen, laces), textural patterns, and hair waves. In particular, our method mitigates the gender bias issue reported in~\cite{DeformableGAN} caused by imbalanced gender distribution in fashion images. As shown in the third row, without explicit facial guidance in the source image, all the other methods tend to predict the more frequently occurred female face by mistake. In contrast, our method is more effective in capturing the fine-grained semantic details, and is able to faithfully retain the masculinity of synthesized person in both coarse and refined outputs.

\textbf{Perceptual Evaluation} As reported in Table \ref{tab:compare_sota}, our DRN achieves significant improvements against recent state-of-the-art methods in terms of perceptual quality, with $5.2$ gain on FID against PATN and $9.8$ gain on DSC. For the LPIPS score, the gains are $0.35$ and $0.15$, respectively. Also, the patchwise signal statistical measure SSIM is comparable. It is noteworthy that although our method has the lowest Inception Score, the qualitative improvement is undeniable, thus highlighting the superiority of proposed perceptual metrics against the unsupervised IS metric.

\begin{table}[!t]
	\renewcommand{\arraystretch}{1.2}
	\centering
	\caption{Performance against other baseline methods, the best and the second best result for each metric is shown in bold / underline format. Up arrow means higher score is preferred, and vice versa.}
	\label{tab:compare_sota}
	\begin{tabular}{|c|cccc|}
		\hline
		Methods & SSIM $ \uparrow $ & IS $ \uparrow $ & FID $ \downarrow $ & LPIPS $ \downarrow $ \\
		\hline
		UV-Net    & 0.763 & \underbar{3.440} & --- & --- \\
		SPT       & 0.736 & \textbf{3.441} & --- & --- \\
		DSC   & 0.762 & 3.330 & 24.479 & \underbar{0.233} \\
		PATN        & 0.773 & 3.209 & 19.816 & 0.253 \\
		\hline
		Ours Coarse        & \textbf{0.780} & 3.230 & \underbar{18.405} & 0.243\\
		Ours Full        & \underbar{0.774} & 3.125 & \textbf{14.611} & \textbf{0.218}\\
		\hline
	\end{tabular}
\end{table}

We also report the scores tested on coarse estimations $I_c$ without the residual map, as shown in Table~\ref{tab:compare_sota}. Once again, SSIM and IS scores show limited reliability as perceptual evaluators, as they both slightly drop after detail replenishment despite the clear perceptual improvement (20\% gain on FID and 10\% gain on LPIPS). Furthermore, even without detail replenishment, our pose transfer branch still achieves better perceptual scores than PATN baseline. From Fig.~\ref{fig:sota}, we can observe that PATN produces too much irregular, self-repetitive patterns, which looks particularly ``fake'' to human eyes. This could possibly explain the superiority of our coarse model over PATN on perceptual scores. In other words, compared to lacking details, it could be even worse to have fake details in the output image --- this again confirms our motivation to suppress detail replenishment in the coarse stage.

\begin{table}[!t]
	\renewcommand{\arraystretch}{1.2}
	\centering
	\caption{Mean distance between feature embeddings and the percentage of identity-preserving cases with images generated from different baselines.}	\label{tab:face_id}
	\begin{tabular}{|c|ccc|}
		\hline
		Metric & L2 error $\downarrow $ & Acc.($\epsilon=0.6$) $\uparrow$ & Acc. ($\epsilon=0.7$) $ \uparrow $ \\
		\hline
		DSC          & 0.720 & 0.078 & 0.383 \\
		PATN     & 0.706 & 0.085 & 0.485  \\
		Ours (w/o face)  & 0.647 & 0.247 & 0.752 \\
		Ours Full        & \textbf{0.615} & \textbf{0.420} & \textbf{0.873}\\
		\hline
	\end{tabular}
\end{table}

\begin{figure}[!t]
	\centering
	\begin{minipage}{\linewidth}
		\includegraphics[width=0.46\linewidth]{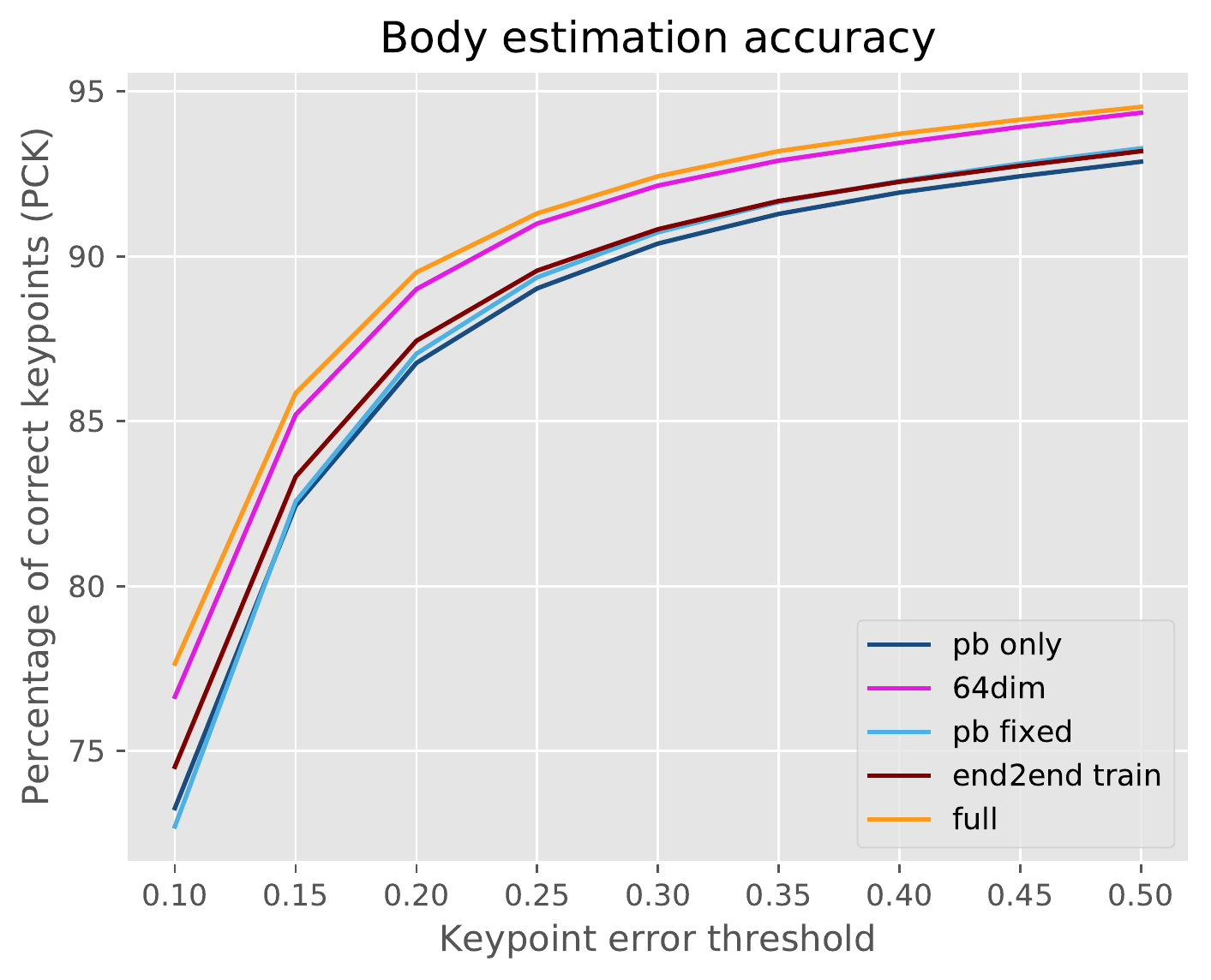}\quad
		\includegraphics[width=0.46\linewidth]{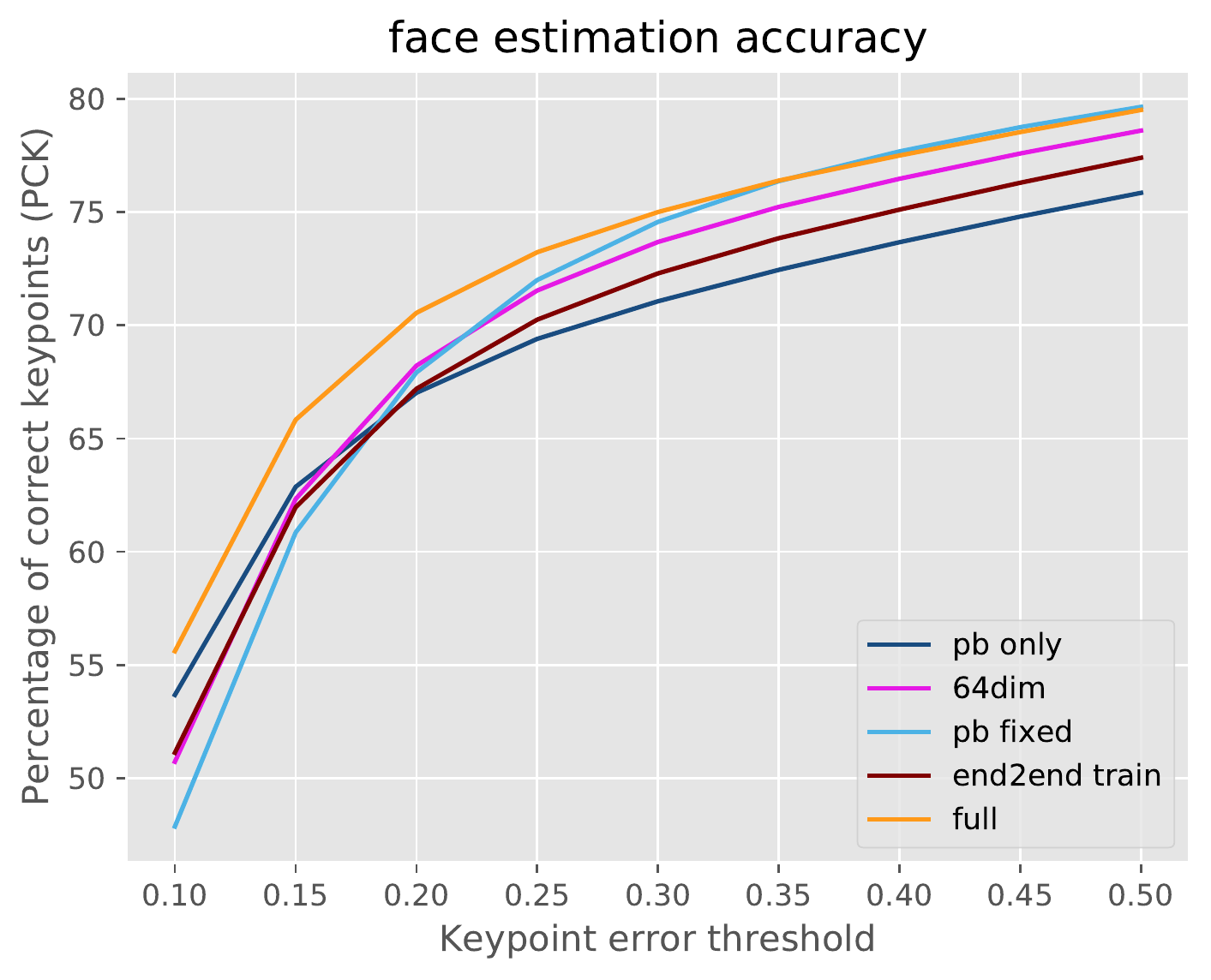}\\
		\includegraphics[width=0.46\linewidth]{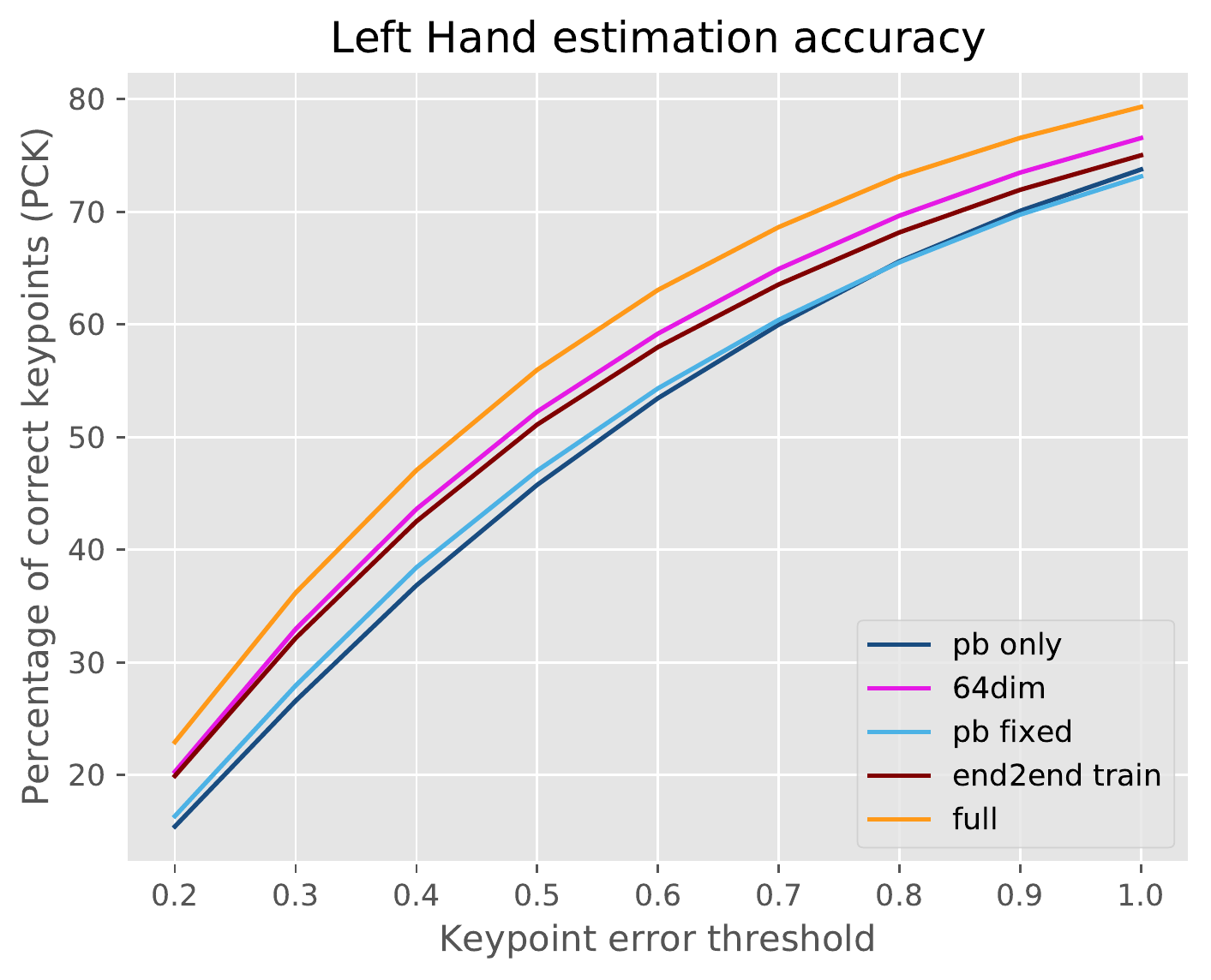}\quad
		\includegraphics[width=0.46\linewidth]{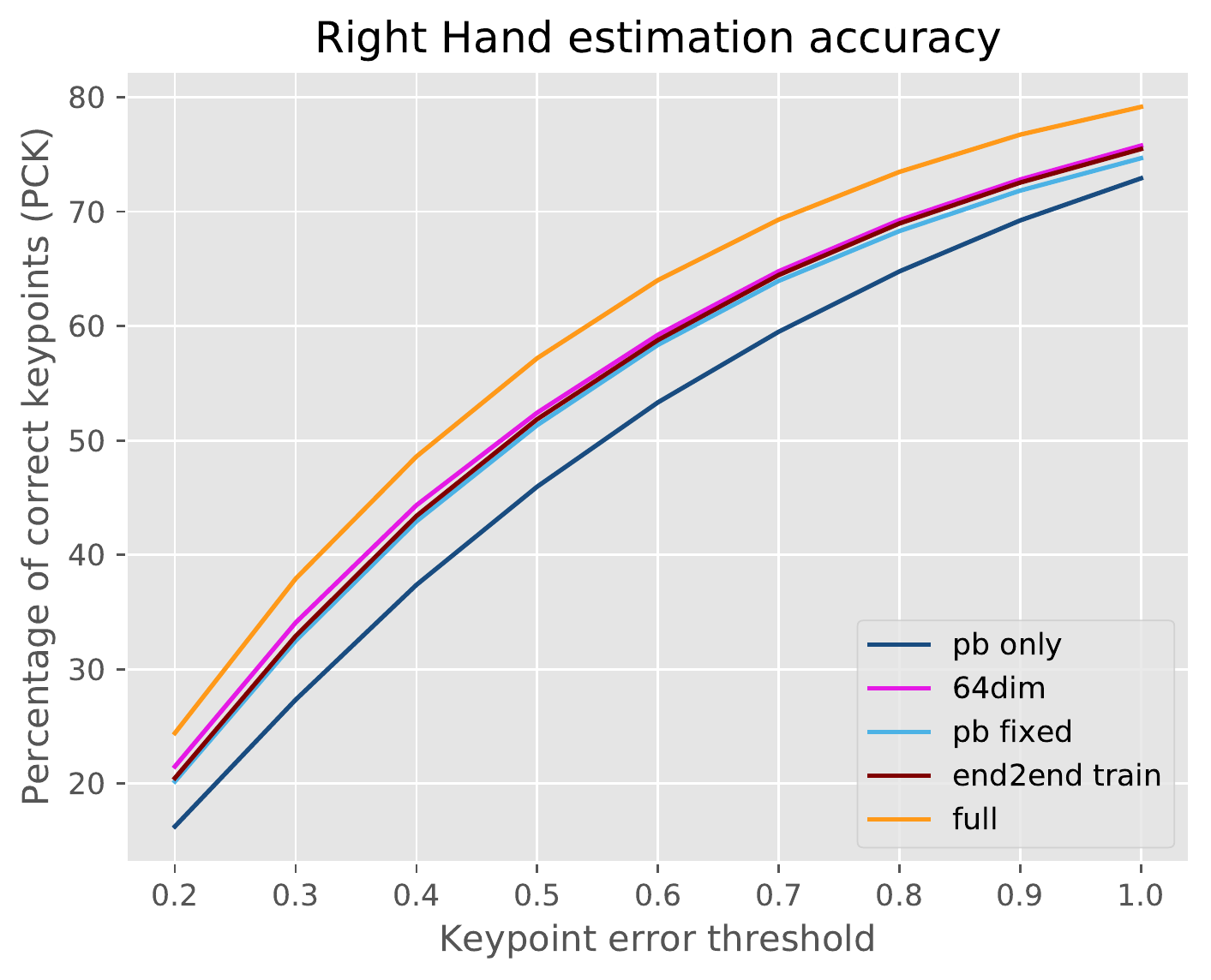}\\
	\end{minipage}
	\caption{Keypoint error curves of ablation methods for each body part. Note that the error levels for both hands are doubled to cover the majority of estimated points. \emph{Best view in color.}}
	\label{fig:keypoint_error_curve}
	
\end{figure}

\textbf{Structural Evaluation}
As shown in Fig.~\ref{fig:keypoint_error_curve}, our DRN has demonstrated consistent quality improvement over the entire human body. Among different body parts, DRN is especially powerful for restoring facial details even without using ground truth facial landmarks. For the body and both hands, the accuracy curves of different methods are roughly parallel, with a gap of 5\% between the proposed DRN and the original PATNetwork (PB only). This indicates that our mutual-guidance design helps refine the structural integrity of generated body parts, thus reducing 5\% outliers with large localization errors. In summary, the proposed DRN not only restores convincing appearance details but also helps improve the structural integrity of body parts.

\begin{figure}[!t]
	\centering
	\includegraphics[width=\linewidth]{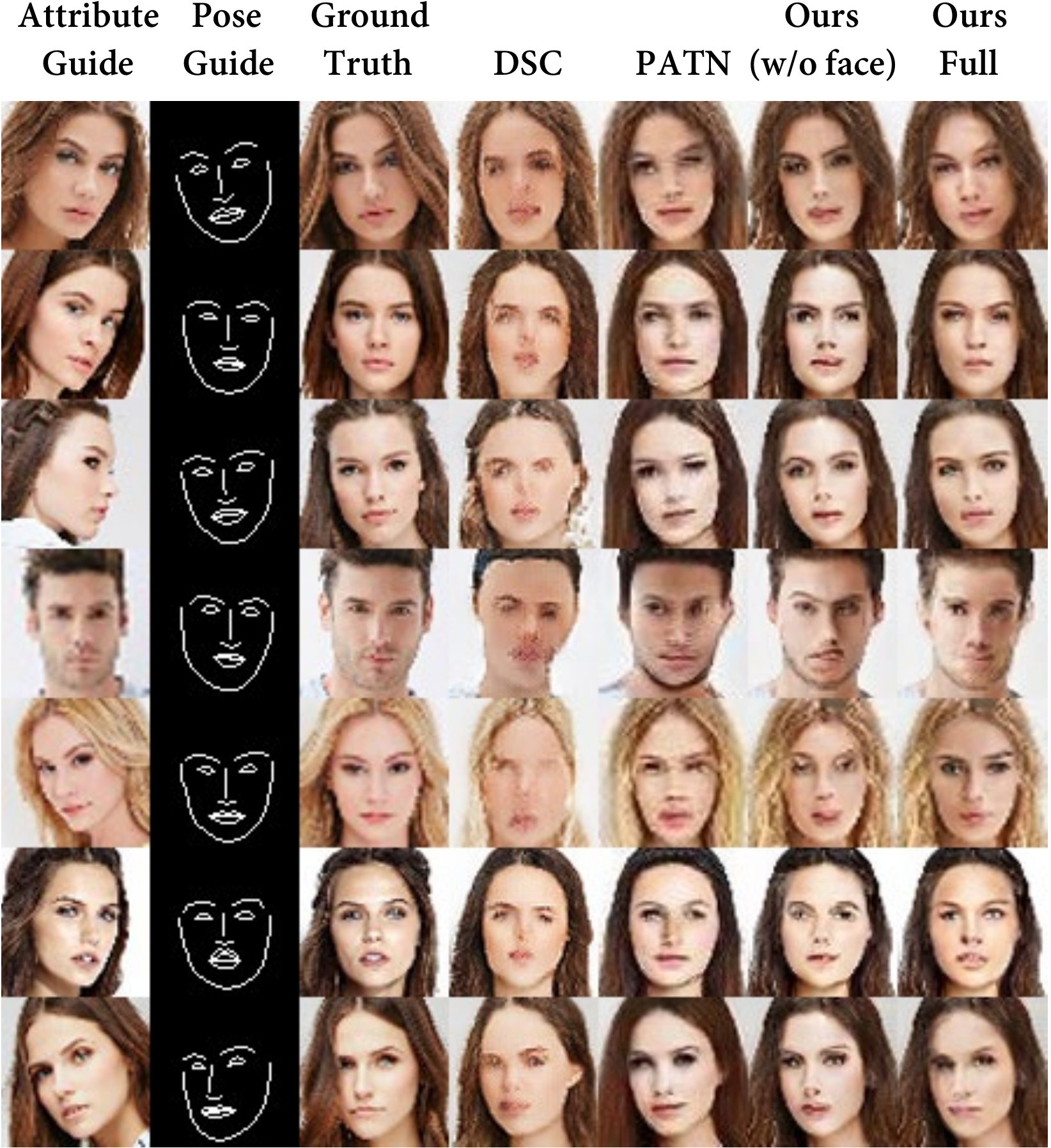}
	\caption{Qualitative results of face attribute transfer with different methods. Our face enhancing module can more faithfully preserve the visual attributes of input faces and prevent distortions. \emph{Please zoom in for details.}}\label{fig:face_ablation}
\end{figure}

\textbf{Semantic Evaluation}\label{sec:face}
We first evaluate the facial identity preservation of different baselines in Table~\ref{tab:face_id} by measuring the mean embedding distance and the corresponding percentage of identity-preserving cases. Our basic framework with global DRM is already leading existing methods by 27\%, with an additional 12\% gain with additional face detail replenishment. We also present the qualitative results in Fig.~\ref{fig:face_ablation}, where DSC~\cite{DeformableGAN} fails to preserve the facial structure, and incurs severe landmark distortions. PATN~\cite{PATN} manages to synthesize more plausible faces, but with limited diversity in hairstyles and lip shapes, and sometimes with false attributes, like the thick black beard for the male in row 4. In contrast, our method can well preserve prominent facial attributes and correct mild distortions, leading to better identity consistency and visual realism of generated faces.

\begin{figure*}[p]
	\centering
	\subfloat[The ``easy'' case]{\includegraphics[width=0.8\textwidth]{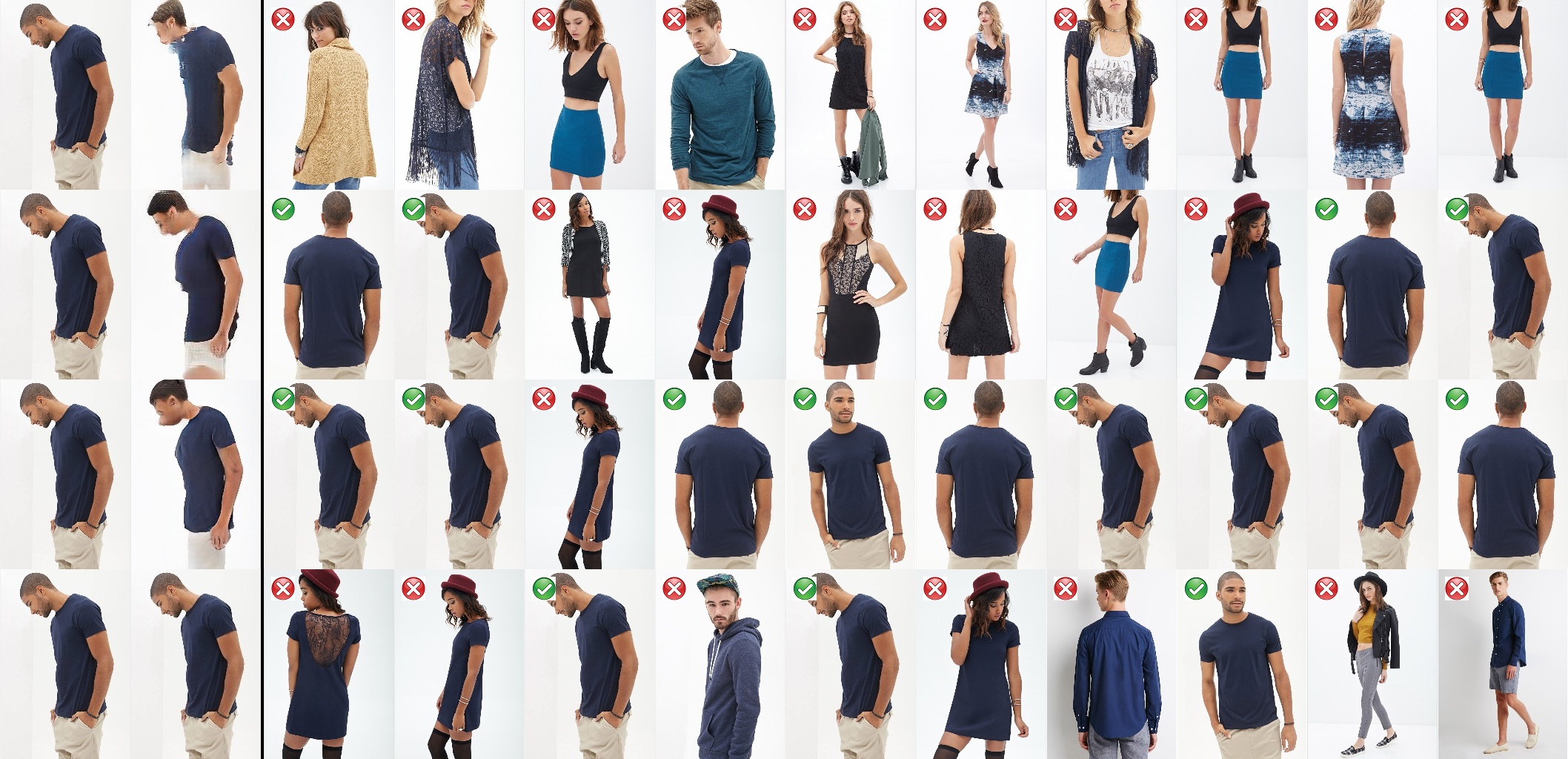}%
		\label{a}}
	\hfil
	\subfloat[The ``challenging'' case]{\includegraphics[width=0.8\textwidth]{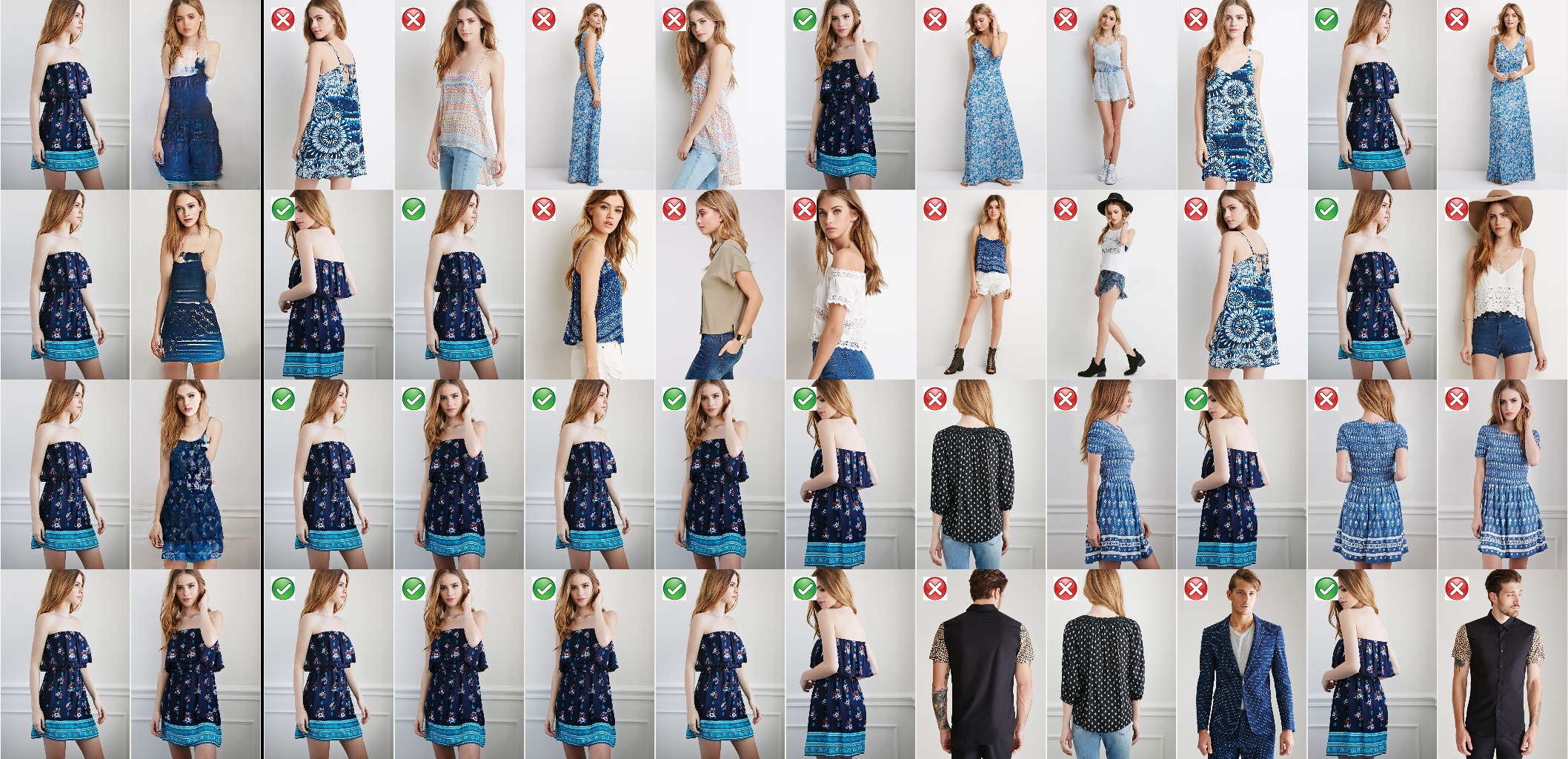}%
		\label{b}}
	\hfil
	\subfloat[The ``confusing'' case]{\includegraphics[width=0.8\textwidth]{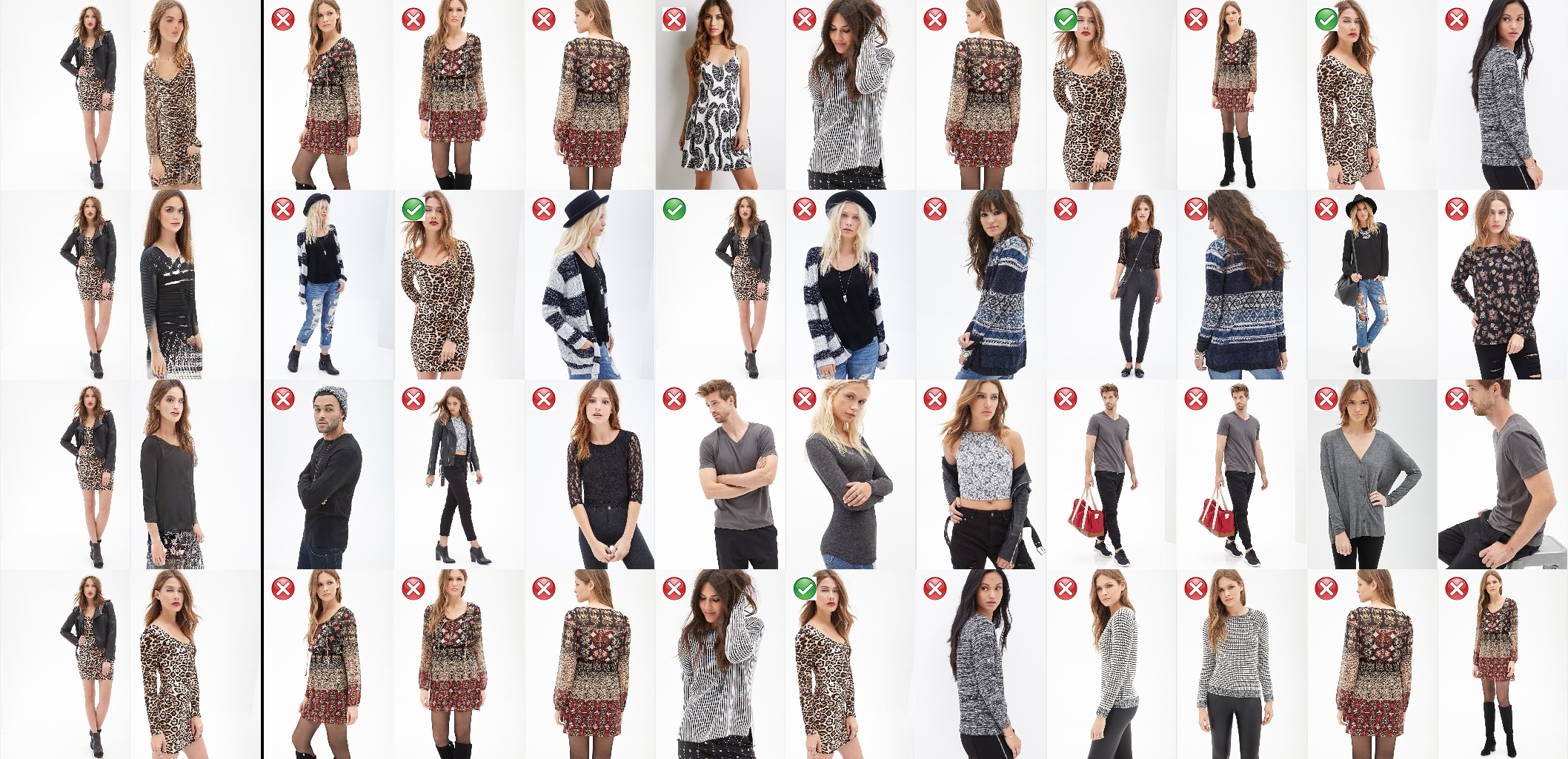}%
		\label{c}}
	\caption{Top 10 matching candidates for image retrieval evaluation. Queries are generated with different baselines. From left to right: Input source, generated query, and retrieval results. From top to bottom: DSC~\cite{DeformableGAN}, PATN~\cite{PATN}, the proposed DRN, and the ground truth. Positive and negative items are marked in green checks and red crosses respectively. \emph{Zoom in for details.}}
	\label{fig:retrieval}
\end{figure*}

\begin{table}[!t]
	\renewcommand{\arraystretch}{1.2}
	\centering
	\caption{Top-K recall performance on in-shop clothes retrieval task with images generated from different baselines.}
	\label{tab:retrieval}
	\begin{tabular}{|p{0.08\textwidth}|p{0.04\textwidth}p{0.04\textwidth}p{0.043\textwidth}|p{0.04\textwidth}p{0.04\textwidth}p{0.04\textwidth}|}
		\hline
		Recall(\%) & &  Resnet50  & & & VGG16 & \\
		& Top 3 & Top 5 & Top~10 & Top 3 & Top 5 & Top~10 \\
		\hline
		PATN & 6.58 & 10.32 & 17.84 & 6.43 & 9.97 & 16.97\\
		DSC & 11.09 & 17.60 & 30.07 & 11.31 & 17.52 & 28.93 \\
		\hline
		PB Only       & 6.56 & 10.23 & 17.78 & 6.35 & 10.02 & 17.10\\
		PB Fixed      & 4.58 & 7.27 & 12.91 & 4.85 & 7.64 & 12.92\\
		Tex-64dim  & 6.59 & 10.32 & 17.80 & 6.95 & 11.06 & 18.06 \\
		End2end       & 6.89 & 10.90 & 18.67 & 6.29 & 9.47 & 16.33\\
		Ours Full     & \textbf{11.79} & \textbf{18.45} & \textbf{30.79} & \textbf{11.61} & \textbf{18.60} & \textbf{30.23}\\
		\hline
		Real Images & 26.13 & 41.36 & 67.89 & 25.08 & 38.83 & 62.92 \\
		\hline
	\end{tabular}
\end{table}

Finally, we evaluate the overall content preservation of our proposed method with the newly proposed content-based image retrieval task.
Table~\ref{tab:retrieval} presents the Top-K recall of different methods for K = 3, 5, 10 under two different backbones, ResNet and VGG. Our DRN has a significant advantage over state-of-the-art PATN baseline, with almost twice the recall rate for every K, thus proves the efficacy of our texture enhancing module for recovering appearance details lost during the pose transfer stage. Also, it is evident that the proposed network architecture and training strategy are complementary with each other, as changing either element would severely degrade the overall performance, leading to barely comparable recall rate against the PATN baseline.

Also, we point out two important observations in Table~\ref{tab:retrieval}. Firstly, the warping-based DSC method significantly outperforms PATN, a recently published work with better visual quality, Fig.~\ref{fig:sota}; Secondly, although our method achieves the top recall performance, it's still far less accurate than querying with real images. To better help with the understanding of the information carried within extracted feature embeddings, and the similarity criterion used by the retrieval system, we compare the top 10 best matching candidates retrieved from queries generated with different methods.

Three typical examples are shown in Fig.~\ref{fig:retrieval}, where (a) represents the easy case with input clothes containing few textures, (b) represents the challenging case with clothes of rich textures, and (c) represents the confusing case with two or more items in different textural styles --- a deep gray coat and a leopard-spotted short skirt.
In case (a), we note the retrieval system frequently mismatches a woman in a deep blue skirt, suggesting a higher focus on color consistency over other attributes such as gender and accessories.
For case (b), the retrieval results are closer to our expectations, where the number of positive matches are growing consistently with the increasing amount of textural detail in query images.
For the confusing case (c), DSC successfully recovers the leopard-spotted skirt from the source image, leading to a more accurate retrieval result with correct top-3 candidates, although not positive matches. In contrast, both our DRN and the PATN baseline failed to capture the desired outfit. This could partially explain DSC's superior performance and help motivate novel hybrid architectures combining the strengths of warping-based and synthesizing-based methods.

In conclusion, our retrieval-based evaluation protocol is more reliable in measuring semantic and appearance consistency of generated human images, especially for the case with rich textures. Also, it is much more flexible in adjusting the similarity criterion according to different attributes, such as colors, fabrics, textures, and hairstyles. We believe that the advances in fashion retrieval systems could help facilitate the design of better similarity criteria that would further boost the visual realism of human-centered image synthesis tasks.

\subsection{Discussion}
\begin{figure}
  \centering
  \includegraphics[width=\linewidth]{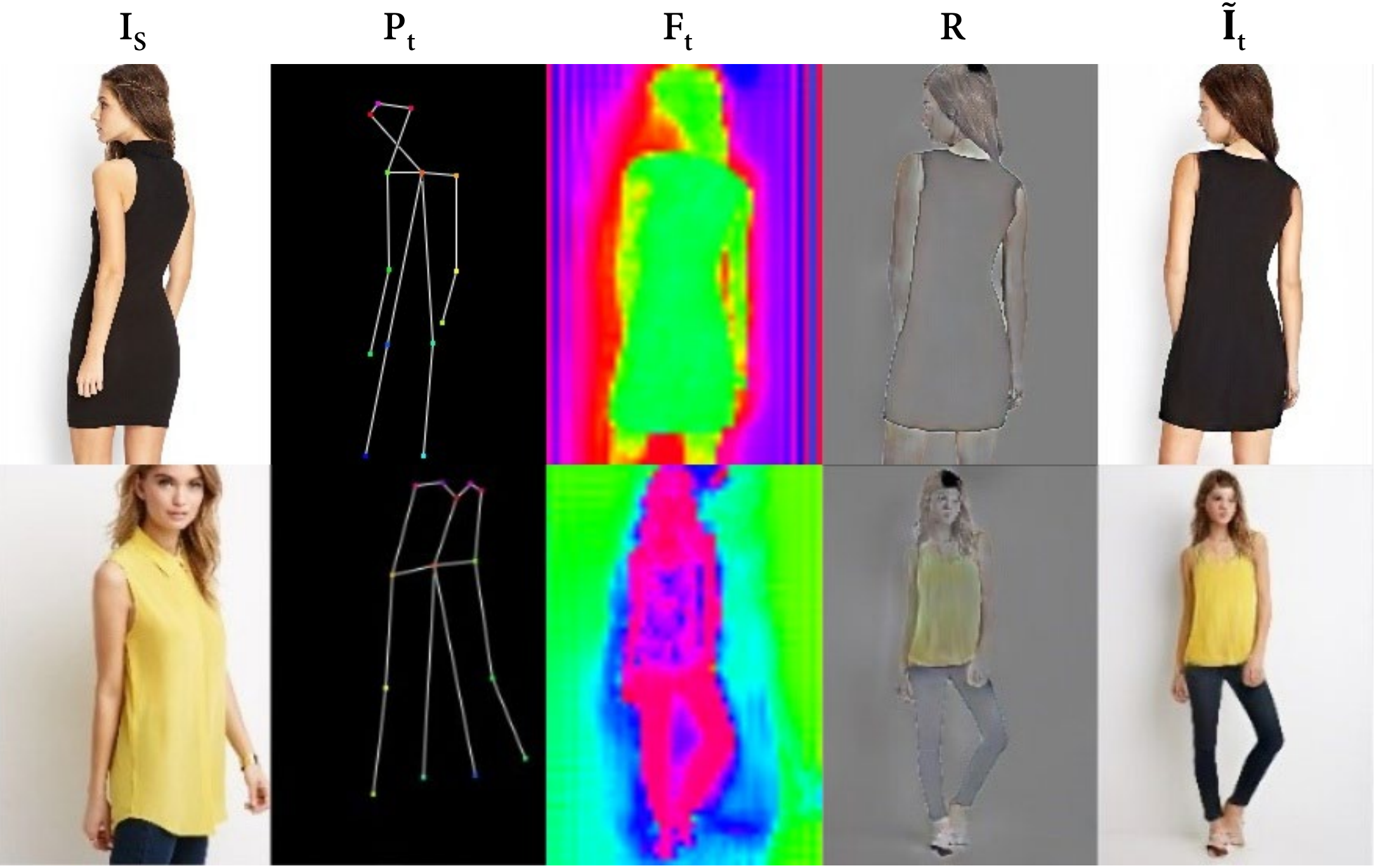}
  \caption{Visualization of semantic guidance map $F_t$}\label{fig:vis}
\end{figure}

\textbf{Semantic Guidance Visualization.} To reveal the working mechanism of semantic-guided detail replenishment, we apply PCA to the guidance map $F_t$ and normalize the first 3 principle channels as RGB values, leading to the visualization result in Fig.~\ref{fig:vis}. Our semantic guidance map is structurally aligned with the target pose $P_t$, thus help replenishing details at corresponding semantic regions.

\begin{table}[!t]
	\renewcommand{\arraystretch}{1.2}
	\centering
	\caption{Gender prediction error rate of different methods.}
	\label{tab:gender_bias}
\begin{tabular}{|c|c|c|c|}
  \hline
  Methods & DSC & PATN & Ours \\
  \hline
  Error rate (\%) & 17.7 & 10.4 & 6.5 \\
  \hline
\end{tabular}
\end{table}

\textbf{Gender Bias.} The DeepFashion benchmark dataset features an uneven gender distribution (7248 male vs 41426 female), which often causes ``gender bias'' ~\cite{DeformableGAN} in existing HPT methods to falsely producing female faces against a male input. To quantify this issue, we train a VGG network for binary gender classification over the DeepFashion training split, and use it to predict the gender of synthesized person images. Table~\ref{tab:gender_bias} shows the error rate of different methods, where our DRN achieves significant bias reduction against other baselines. One possible explanation is that our detail replenishment module is able to capture the contextual relationship between gender and facial attributes/garment styles with the attribute vector $z_s$. For instance, a man is more likely to wear T-shirts, grow beards, and have tattoos in the arm. We shall note, however, that so far this is still a hypothesis without solid experimental proof, yet it provides an incentive to further explore the causality between gender and visual attributes, which can hopefully lead to better gender equality in future FHPT solutions.

\section{Conclusions}

Human pose transfer (HPT) is an emerging research topic with huge application potential in creative media applications. Yet current HPT methods typically introduce \emph{detail deficiency, content ambiguity} or \emph{style inconsistency} in synthesized person images due to the suboptimal integration between low-level feature transfer and high-level semantic-guided content synthesis. To address existing issues and narrow the theory-practice gap of HPT, we aim towards a more practical and challenging setting, termed as Fine-Grained Human Pose Transfer (FHPT). Specifically, we establish the new objectives for FHPT task: \emph{perceptual realism, structural integrity, and semantic fidelity}, with a comprehensive suite of fine-grained evaluation protocols targeted at these objectives, including face identity preservation, keypoint localization, and content-based image retrieval,  providing a comprehensive, accurate and reliable measurement of the model capability towards FHPT scenarios.
To implement FHPT, we propose a Detail Replenishing Network (DRN) with distinctive structural designs, including a guided detail replenishment branch for style-consistency in generated contents, and an intermediate feature-sharing path to facilitate the mutual guidance between the global predictive and local warping branches. As a result, DRN achieve significant performance gain against varying HPT baselines by a huge margin, highlighting the challenges in the FHPT task as well as the efficacy of our contributions.

In the future, we aim to investigate more effective semantic guidance and feature utilization schemes for FHPT. For instance, it would be beneficial to incorporate region adaptive style control into the FHPT framework. Also, designing interpretable and customizable evaluation protocols is another promising direction. Finally, although our framework is designed for synthesizing human images, we note that the proposed ``detail-replenishment'' approach can also benefit other semantic-guided image synthesis tasks, such as face renovation~\cite{HiFaceGAN}, leading to enhanced appearance quality and perceptual realism in produced contents.

\ifCLASSOPTIONcaptionsoff
  \newpage
\fi



\bibliographystyle{IEEEtranS}
\bibliography{icme2020blind}
%
%
%

%
\begin{IEEEbiography}
[{\includegraphics[width=1in,height=1.25in,clip,keepaspectratio]{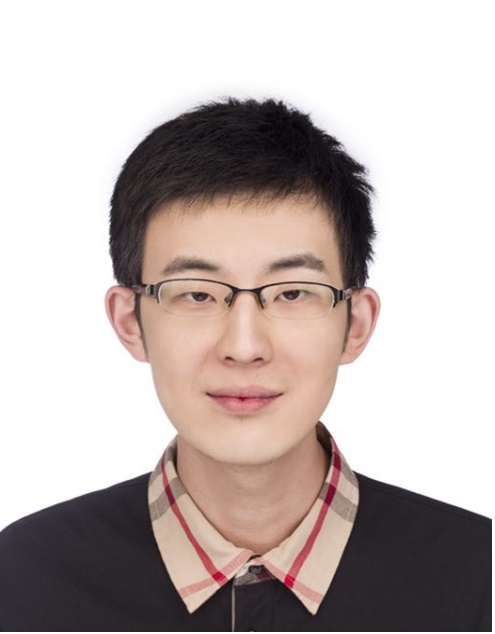}}]{Lingbo Yang} received the BS degree in mathematics
and applied mathematics from Peking University in
2016. He is currently pursuing the PhD degree at
Institute of Digital Media, Peking University. He has
been interning at DAMO Academy, Alibaba Group
since 2019. His research interests include deep gen-
erative models, image restoration and editing, and
human pose transfer. In 2020, he has authored 6 peer-reviewed journals and conference papers, yet his best paper is always the next one.
\end{IEEEbiography}

\begin{IEEEbiography}
[{\includegraphics[width=1in,height=1.25in,clip,keepaspectratio]{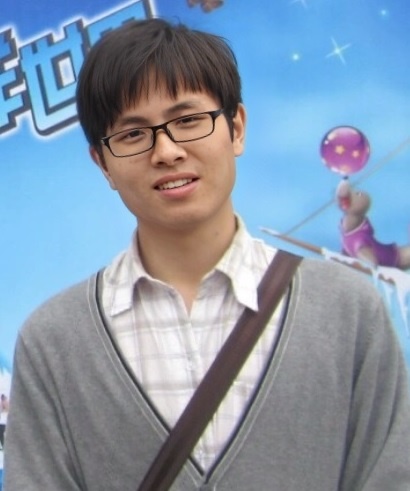}}]{Pan Wang}
received the B.S. degree in information
and control engineering from China University of
Petroleum, Qingdao, China, in 2013, and the M.S.
degree in computer science with the School of
Electronic, Electrical and Communication Engineer-
ing, University of Chinese Academy of Sciences,
Beijing, China, in 2018. He is now a computer vision
algorithm engineer at Alibaba DAMO academy. His
current research interests include deep generative
models, image restoration, video inpainting and ob-
ject tracking.
\end{IEEEbiography}
\begin{IEEEbiography}
[{\includegraphics[width=1in,height=1.25in,clip,keepaspectratio]{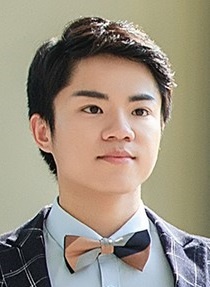}}]{Chang Liu}
 received the B.S. degree from Jilin
University, Jilin, China, in 2012. Since 2015, he
has been a Ph.D student in the School of Elec-
tronic, Electrical and Communication Engineering,
University of Chinese Academy of Sciences, Bei-
jing, China. His research interests include computer
vision and machine learning, specifically for neural
architecture design and visual object detection. He
has published more than 10 papers in referred con-
ferences including ECCV, ICCV and CVPR.
\end{IEEEbiography}
\begin{IEEEbiography}
[{\includegraphics[width=1in,height=1.25in,clip,keepaspectratio]{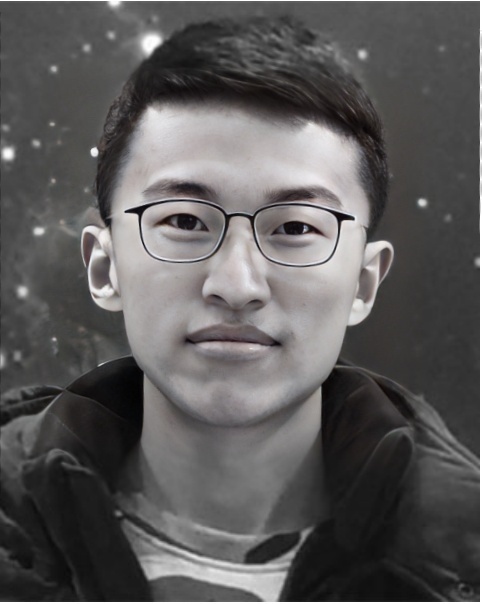}}]{Zhanning Gao}
 received the BS degree in automatic
control engineering from Xi’an Jiaotong University,
Xi’an, China, in 2012. He is currently working
toward the PhD degree in the Institute of Artificial
Intelligence and Robtics, Xi’an Jiaotong University.
He was a research intern in Visual Computing
Group, Microsoft Research Asia from 2015 to 2017.
His research interests include compact image/video
representation, large scale content based multimedia
retrieval, and complex event video analysis.
\end{IEEEbiography}
\begin{IEEEbiography}
[{\includegraphics[width=1in,height=1.25in,clip,keepaspectratio]{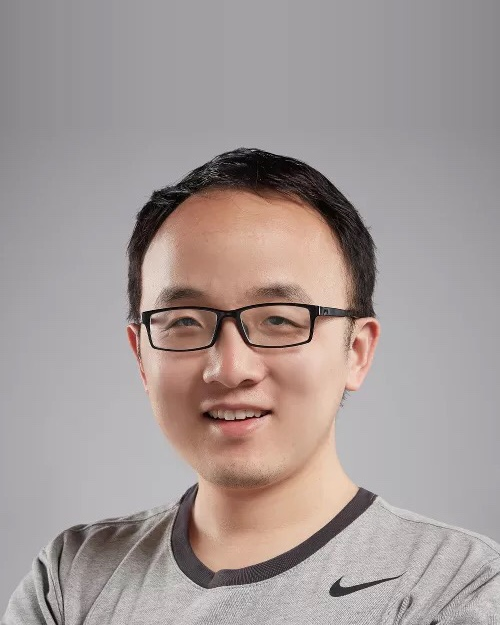}}]{Peiran Ren}
 received his BSc and PhD degree
from Tsinghua University, China, in 2008 and 2014
respectively. He is now a senior algorithm engineer
at Alibaba Damo Acadamy. His research interests
include image and video enhancement and process-
ing, computer aided design, real-time rendering, and
appearance acquisition.
\end{IEEEbiography}
\begin{IEEEbiography}
[{\includegraphics[width=1in,height=1.25in,clip,keepaspectratio]{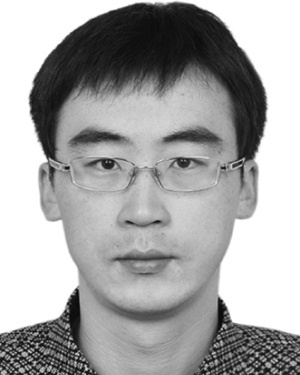}}]{Xinfeng Zhang}
received the BSc from the Hebei University of Technology in 2007, and PhD degree from Chinese Academy of Sciences in 2014. He served as research fellow/postdoc in Nanyang Technological University, University of Southern California, and City University of Hong Kong. He is currently an
Assistant Professor with the Department of Computer Science, University
of Chinese Academy of Sciences. He authored 20 technical
proposals to ISO/MPEG, ITU-T, and AVS standards and more
than 100 refereed journals/conference papers. His research interests include
video compression, image/video quality assessment, and image/video analysis.
He received the Best Paper Award at the 2017 Pacific-Rim Conference on
Multimedia, the Best Paper Award of IEEE Multimedia 2018, and is the
coauthor of a paper that received the Best Student Paper Award in the IEEE
International Conference on Image Processing 2018.
\end{IEEEbiography}
\begin{IEEEbiography}
[{\includegraphics[width=1in,height=1.25in,clip,keepaspectratio]{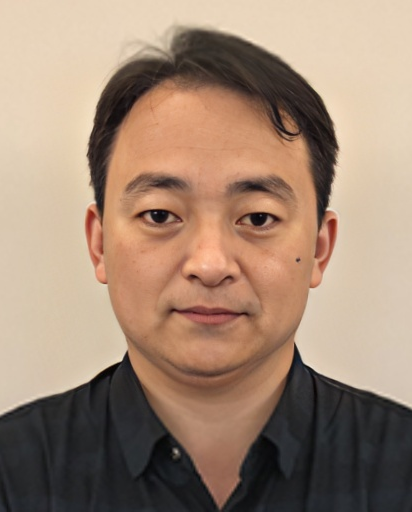}}]{Shanshe Wang}
 received the B.S. degree from the
Department of Mathematics, Heilongjiang Univer-
sity, Harbin, China, in 2004, the M.S. degree in com-
puter software and theory from Northeast Petroleum
University, Daqing, China, in 2010, and the Ph.D.
degree in computer science from the Harbin Institute
of Technology. He held a post-doctoral position with
Peking University from 2016 to 2018. He joined
the School of Electronics Engineering and Com-
puter Science, Institute of Digital Media, Peking
University, Beijing, where he is currently a Research
Assistant Professor. His current research interests include video compression
and image and video quality assessment.
\end{IEEEbiography}
\begin{IEEEbiography}
[{\includegraphics[width=1in,height=1.25in,clip,keepaspectratio]{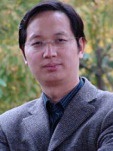}}]{Xiansheng Hua}
received the B.S. and Ph.D. degrees
in applied mathematics from Peking University, Bei-
jing, in 1996 and 2001, respectively. In 2001, he
joined Microsoft Research Asia as a Researcher.
He became a Researcher and the Senior Director
of the Alibaba Group in 2015. He has authored
or co-authored over 250 research papers and has
filed over 90 patents. His research interests have
been in the areas of multimedia search, advertising,
understanding, and mining, and pattern recognition
and machine learning. He was honored as one of the
recipients of MIT35. He served as a Program Co-Chair for the IEEE ICME
2013, the ACM Multimedia 2012, and the IEEE ICME 2012, and on the
Technical Directions Board of the IEEE Signal Processing Society. He is an
ACM Distinguished Scientist and IEEE Fellow.
\end{IEEEbiography}
\begin{IEEEbiography}
[{\includegraphics[width=1in,height=1.25in,clip,keepaspectratio]{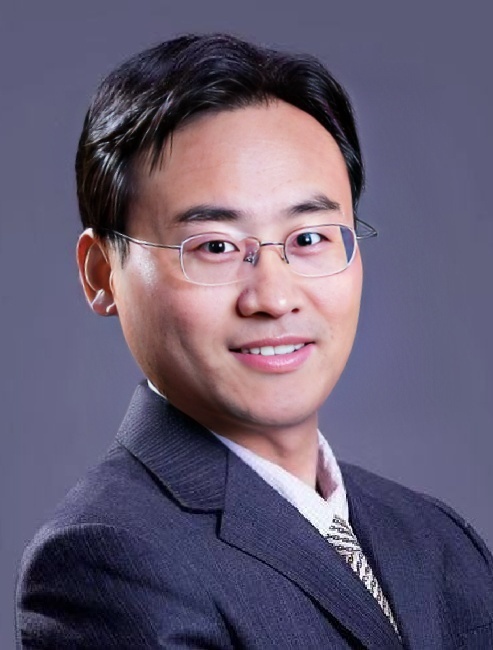}}]{Siwei Ma}
(Senior Member, IEEE) received the BSc
degree from Shandong Normal University in 1999, and the Ph.D. in computer
science from the Institute of Computing Technology,
Chinese Academy of Sciences in
2005. He worked as a postdoc with the
University of Southern California from 2005 to 2007. He joined the
Institute of Digital Media, Peking University where he is currently a Professor. He has
authored over 200 technical articles in refereed journals and proceedings in
image and video coding, video processing, video streaming, and transmission.
He is an Associate Editor of the IEEE TRANSACTIONS ON CIRCUITS
AND SYSTEMS FOR VIDEO TECHNOLOGY and the Journal of Visual
Communication and Image Representation.
\end{IEEEbiography}
\begin{IEEEbiography}
[{\includegraphics[width=1in,height=1.25in,clip,keepaspectratio]{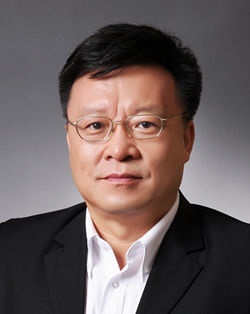}}]{Wen Gao}
 (Fellow, IEEE) received the Ph.D. de-
gree in electronic engineering from The University
of Tokyo in 1991. He was a Professor
of computer science with the Harbin Institute of
Technology from 1991 to 1995 and with the Institute of Computing Technology, Chinese Academy of Sciences, from 1996 to 2006.
He is currently a Professor of computer science
with Peking University. He has authored extensively, including five books and more
than 600 technical articles in refereed journals and
conference proceedings in the areas of image processing, video coding
and communication, pattern recognition, multimedia information retrieval,
multimodal interface, and bioinformatics. He chaired a number of prestigious
international conferences on multimedia and video signal processing, such
as IEEE ISCAS, ICME, and the ACM Multimedia, and also served on the
advisory and technical committees for numerous professional organizations.
He served or serves on the Editorial Board for several journals, including TCSVT, TMM, TIP, TAMD, etc.
\end{IEEEbiography}

%
%
%




\end{document}